\documentclass[letterpaper, 10 pt, journal, twoside]{IEEEtran}
\usepackage[utf8]{inputenc} 
\usepackage[T1]{fontenc}    
\usepackage{url}            
\usepackage{booktabs}       
\usepackage{amsfonts}       
\usepackage{nicefrac}       
\usepackage{microtype}      
\usepackage{xcolor}         

\usepackage{anyfontsize}
\usepackage{amsmath}
\usepackage{amssymb}
\usepackage{wrapfig}
\usepackage{subfig}
\usepackage{caption}
\usepackage{color}

\usepackage{physics}
\usepackage{bm}

\usepackage{algorithm}
\usepackage{algorithmic}
\usepackage{graphicx}

\usepackage{ragged2e}
\usepackage{fp}

\usepackage{ifthen}

\usepackage{pgfplots}
\usepackage{tikz}
\usepackage{tikz-3dplot}

\usepackage{ragged2e}
\usepackage{fp}

\usepackage{ifthen}

\usepackage{tikz}
\usetikzlibrary{
shapes.geometric, 
arrows,
 calc,
 automata,
 positioning,
 decorations,
 decorations.text,
 patterns}
 
 \usetikzlibrary{pgfplots.statistics, pgfplots.colorbrewer} 
\usepackage{pgfplotstable}

\usetikzlibrary{patterns.meta}

\usetikzlibrary{mindmap,snakes}

\usepackage{multirow}

\usepackage{algorithm}
\usepackage{algorithmic}
\usepackage{graphicx}
\usepackage[english]{babel}
\usepackage{array}
\usepackage[export]{adjustbox}

\usepackage{nicefrac}

\usepackage{xcolor}

\usepackage{colortbl}
\usepackage{makecell}
\usepackage{pifont}

\usepackage{mathptmx}

\usepackage[colorlinks=true,
    linkcolor=red,
    linkbordercolor=red,
    urlcolor=mylinkcolor,
    urlbordercolor=white
    ]{hyperref}

\definecolor{mylinkcolor}{rgb}{0.005, 0.3, 0.7}
\definecolor{suppcolor}{rgb}{0.6, 0.0, 0.9}

\newcommand{\OURS}{TMDC}
\newcommand{\OURSFull}{Thrust Microstepped Decoupled Control}

\newcommand{\cfontpt}[1]{\fontsize{5.7}{\baselineskip}{\selectfont #1}}
\usepackage[font=footnotesize]{caption}

\title{Thrust Microstepping via Acceleration Feedback in Quadrotor Control for Aerial Grasping of Dynamic Payload \vspace{-0.75ex} }

\author{Ashish~Kumar$^{\dagger}$, Laxmidhar~Behera$^{\dagger}$, \IEEEmembership{Senior Member, IEEE} \vspace{-1.75ex}  \\
\thanks{Manuscript received: November 30, 2023; Accepted December 4, 2023. This paper was recommended for publication by Editor Giuseppe Loianno upon evaluation of the Associate Editor and Reviewers' comments.}
\thanks{$^{\dagger}$EE, Indian Institute of Technology (IIT), Kanpur, India.
{\tt\small \{krashish,lbehera\}@iitk.ac.in}}
\thanks{Digital Object Identifier (DOI): see top of this page.}
}

\begin{document}

\maketitle

\begin{justify}

\begin{abstract}
In this work, we propose an end-to-end Thrust Microstepping and Decoupled Control (\OURS) of quadrotors. \OURS{} focuses on precise off-centered aerial grasping of payloads dynamically, which are attached rigidly to the UAV body via a gripper contrary to the swinging payload. The dynamic payload grasping quickly changes UAV's mass, inertia etc, causing instability while performing a grasping operation in-air. We identify that to handle unknown payload grasping, the role of thrust controller is crucial. Hence, we focus on thrust control without involving system parameters such as mass etc. \OURS{} is based on our novel Thrust Microstepping via Acceleration Feedback (\texttt{TMAF}) thrust controller and Decoupled Motion Control (\texttt{DMC}). \texttt{TMAF} precisely estimates the desired thrust even at smaller loop rates while \texttt{DMC} decouples the horizontal and vertical motion to counteract disturbances in the case of dynamic payloads. We prove the controller's efficacy via exhaustive experiments in practically interesting and adverse real-world cases, such as fully onboard state estimation without any positioning sensor, narrow and indoor flying workspaces with intense wind turbulence, heavy payloads, non-uniform loop rates, etc. Our \OURS{} outperforms recent direct acceleration feedback thrust controller (DA) and geometric tracking control (GT) in flying stably for aerial grasping and achieves RMSE below $0.04$m in contrast to $0.15$m of DA and $0.16$m of GT.
%
%

\end{abstract}

\end{justify}

\vspace{-0.75ex}

\vspace{-0.00ex}
\noindent
\textbf{Code:} \textcolor{mylinkcolor} {\footnotesize \url{https://github.com/ashishkumar822/\OURS}}~~\textbf{Video:}~\textcolor{gray}{\footnotesize See attachment.}

\IEEEpeerreviewmaketitle

\section{Introduction}
\label{sec:intro}
\begin{figure}[t]
\centering
\begin{tikzpicture}

\FPeval{\circwidth}{0.15}
\FPeval{\dotscale}{0.3}
\FPeval{\circscale}{0.5}
\colorlet{dotcolor}{white!0!green}
\colorlet{circcolor}{white!30!orange}
\colorlet{linkcolor}{black!0!orange}
\colorlet{boxdrawcolor}{white!70!orange}
\colorlet{boxfillcolor}{white!90!black}
\FPeval{\txtscale}{0.8}
\node (ws) [xshift = 28.0ex,yshift=0ex]{\includegraphics[scale=0.147]{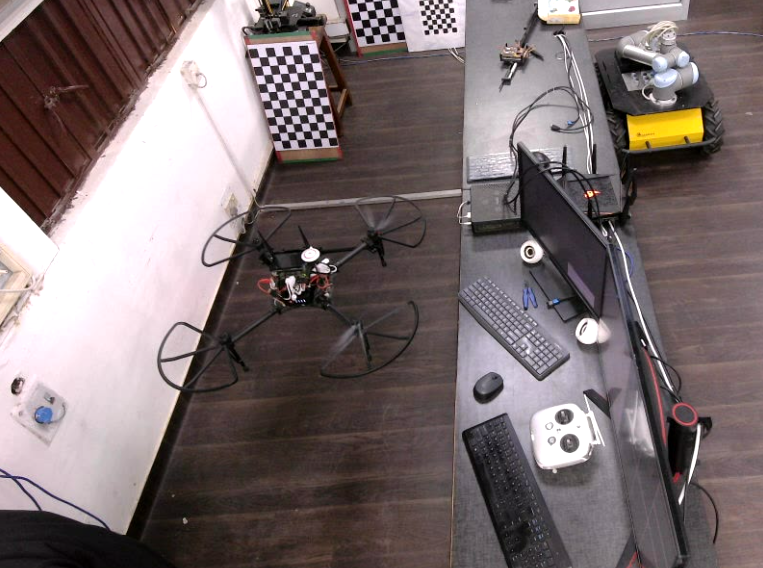}};
\node (m100) [draw=none,xshift=0ex, yshift=0ex,scale=0.6]
{
\tikz{
%
\node (m100_side) [xshift = 0ex,yshift=0ex]{\includegraphics[scale=0.089]{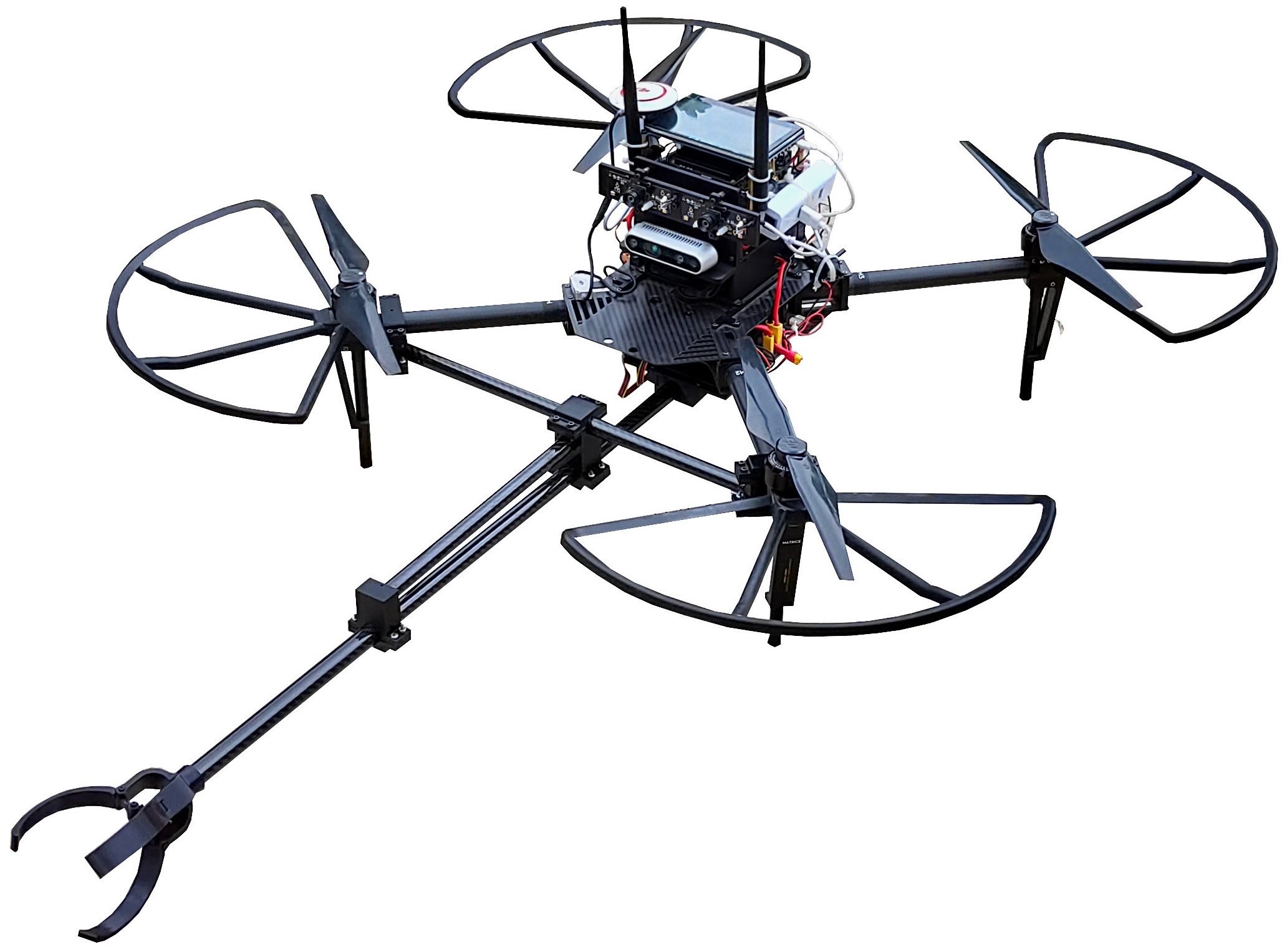}};
%
\node (m600_boundary) [draw=white!60!black,rounded corners=0.5mm,rectangle, align=center,minimum width=47ex,minimum height=31ex,xshift=0ex,yshift=0ex]{};
\node (wifidot) [fill=dotcolor, circle, xshift =-0.5ex,yshift=13.0ex, scale=\dotscale]{};
\node (wifidotcirc) [draw=circcolor, circle, xshift =-0.5ex,yshift=13.0ex, line width=\circwidth ex,scale=\circscale]{};
\node (wifi) [draw=boxdrawcolor,fill=boxfillcolor,rounded corners=0.3mm, rectangle,align=center,xshift =-17.0ex,yshift=13.4ex,scale=\txtscale]{WiFi Antenna};
\draw [-, linkcolor] (wifidotcirc) -- (wifi);
\node (dispdot) [fill=dotcolor, circle, xshift =3.0ex,yshift=11.0ex, scale=\dotscale]{};
\node (dispdotcirc) [draw=circcolor, circle, xshift =3.0ex,yshift=11.0ex, line width=\circwidth ex,scale=\circscale]{};
\node (disp) [draw=boxdrawcolor,fill=boxfillcolor,rounded corners=0.3mm, rectangle,align=center,xshift =19.3ex,yshift=13.4ex,scale=\txtscale]{Display};
\draw [-, linkcolor] (dispdotcirc) -- (disp);
\node (propdot) [fill=dotcolor, circle, xshift =15.0ex,yshift=5.5ex, scale=\dotscale]{};
\node (propdotcirc) [draw=circcolor, circle, xshift =15.0ex,yshift=5.5ex, line width=\circwidth ex,scale=\circscale]{};
\node (prop) [draw=boxdrawcolor,fill=boxfillcolor,rounded corners=0.3mm, rectangle,align=center,xshift =18.5ex,yshift=1.5ex,scale=\txtscale]{Propellers};
\draw [-, linkcolor] (propdotcirc) -- (prop);
\node (stereodot) [fill=dotcolor, circle, xshift =3.0ex,yshift=7.5ex, scale=\dotscale]{};
\node (stereodotcirc) [draw=circcolor, circle, xshift =3.0ex,yshift=7.5ex, line width=\circwidth ex,scale=\circscale]{};
\node (stereo) [draw=boxdrawcolor,fill=boxfillcolor,rounded corners=0.3mm, rectangle,align=center,xshift =18.3ex,yshift=-6.0ex,scale=\txtscale]{Stereo Rig};
\draw [-, linkcolor] (stereodotcirc) -- (stereo);
\node (d435idot) [fill=dotcolor, circle, xshift =-0.4ex,yshift=7.5ex, scale=\dotscale]{};
\node (d435idotcirc) [draw=circcolor, circle, xshift =-0.4ex,yshift=7.5ex, line width=\circwidth ex,scale=\circscale]{};
\node (d435i) [draw=boxdrawcolor,fill=boxfillcolor,rounded corners=0.3mm, rectangle,align=center,xshift =14.8ex,yshift=-13.5ex,scale=\txtscale]{Intel D435i};
\draw [-, linkcolor] (d435idotcirc) -- (d435i);
\node (gripperdot) [fill=dotcolor, circle, xshift =-14.0ex,yshift=-9.0ex, scale=\dotscale]{};
\node (gripperdotcirc) [draw=circcolor, circle, xshift =-14.0ex,yshift=-9.0ex, line width=\circwidth ex,scale=\circscale]{};
\node (gripper) [draw=boxdrawcolor,fill=boxfillcolor,rounded corners=0.3mm, rectangle,align=center,xshift =-9.9ex,yshift=-13.5ex,scale=\txtscale]{Gripper};
\draw [-, linkcolor] (gripperdotcirc) -- (gripper);
\node (armdot) [fill=dotcolor, circle, xshift =-8.8ex,yshift=-4.0ex, scale=\dotscale]{};
\node (armdotcirc) [draw=circcolor, circle, xshift =-8.8ex,yshift=-4.0ex, line width=\circwidth ex,scale=\circscale]{};
\node (arm) [draw=boxdrawcolor,fill=boxfillcolor,rounded corners=0.3mm, rectangle,align=center,xshift =-20.4ex,yshift=-7.0ex,scale=\txtscale]{Arm};
\draw [-, linkcolor] (armdotcirc) -- (arm);
\node (grippersupportdot) [fill=dotcolor, circle, xshift =-1.5ex,yshift=1.5ex, scale=\dotscale]{};
\node (grippersupportdotcirc) [draw=circcolor, circle, xshift =-1.5ex,yshift=1.5ex, line width=\circwidth ex,scale=\circscale]{};
\node (grippersupport) [draw=boxdrawcolor,fill=boxfillcolor,rounded corners=0.3mm, rectangle,align=center,xshift =2ex,yshift=-10.0ex,scale=\txtscale]{Gripper Support};
\draw [-, linkcolor] (grippersupportdotcirc) -- (grippersupport);
\node (jetsondot) [fill=dotcolor, circle, xshift =1.1ex,yshift=10.2ex, scale=\dotscale]{};
\node (jetsondotcirc) [draw=circcolor, circle, xshift =1.1ex,yshift=10.2ex, line width=\circwidth ex,scale=\circscale]{};
\node (jetson) [draw=boxdrawcolor,fill=boxfillcolor,rounded corners=0.3mm, rectangle,align=center,xshift =-18.2ex,yshift=0.0ex,scale=\txtscale]{Jetson NX};
\draw [-, linkcolor] (jetsondotcirc) -- (jetson);
}
};

\end{tikzpicture}
\vspace{-3.0ex}
\caption{Left: Our aerial manipulator. Right: \OURS~controlling a very large sized quadrotor ($1.30$m$\times0.90$m$\times0.45$m) within a workspace clearance of $10$cm at a very small altitude of $0.5$m, a quite adverse and dangerous practice considering the large sized quadrotor.}
\label{fig:m100}
\vspace{-3.3ex}
\end{figure}
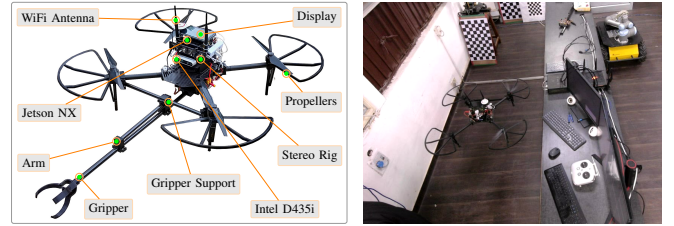

\IEEEPARstart{A}{erial} manipulation is a highly challenging research area and a convoluted design practice. 
It attracts a variety of applications, such as transporting rigidly attached or swinging payloads, gripper-based in-air grasping, etc. This paper focuses on the control of an aerial grasping system for small items, i.e. a quadrotor endowed with a gripper having a three-fingered jaw end-effector (Fig.~\ref{fig:m100}). Notably, developing the control system for such a system is increasingly difficult due to its inherent challenges, as discussed below.
\par
Foremost is the necessity that unlike swinging payloads \cite{xian2020robust}, aerial grasping requires approaching the target laterally. This gives rise to \textit{off-centred} end-effector configuration as it is the natural setup to grasp an item mid-air. This causes a permanent shift in the centre of gravity (CoG). Then after grasping, the grasped item further shifts the CoG dynamically, leading to dynamic payload situations full of critical challenges such as; \textit{First}, according to our experiments, for grasping an item, the UAV needs to approach the target slowly and requires high stability in height as well as in lateral direction. Without that, it is extremely likely to miss the target since the gripper is a small entity, and even a small external disturbance can fail the grasp.
\par
\textit{Second}, due to the fast-changing system dynamics (mass, CoG, etc) during payload attach (grasp) and detach (release) events, the platform is most vulnerable. A centered payload only exerts a downward force, however, in addition to the downward force, an off-center load (grasped item) exerts an intensive torque along the roll or pitch axis depending upon the gripper mounting, thus causing severe unwanted motion of the UAV, resulting in a crash. Notice that this is different from swinging payloads, which are attached beforehand. In contrast, in our case, the payload is attached dynamically autonomously and is held into the gripper tightly. 
\par
Further, from a practical standpoint, dealing with the non-uniform execution rates of the control system is another critical challenge. This challenge arises because, in aerial grasping, several compute-intensive sub-systems co-exist, such as deep neural networks and visual SLAM, which occupy the computing resources intensively, thus preventing uniform loop rates. Since a control system is sensitive to the loop rates, an execution delay of even $5$-$10$ms can degrade its performance \cite{towards}, thus preventing the control system from handling intense disturbances, e.g. torque caused by grasping an off-center item.
\par
Furthermore, precise take-off-hover-land in the presence of dynamic payloads, inaccurate system parameters, imbalanced platform due to the off-center grippers, battery discharging, etc., make the overall problem quite convoluted due to an imprecise thrust estimation. Thus, the control system design for aerial grasping is currently a state-of-the-art problem. 
\par
In this work, we underline that these challenges are closely related to the thrust controller, which is a crucial component of the quadrotor control. Typically, the quadrotor control can be sectioned into an outer loop (position controller) and an inner loop (attitude controller). These loops strictly depend on the system parameters, such as mass and propeller coefficient, and are severely affected by their incorrect values.
\par
The position controller has a position feedback loop that computes acceleration from a desired $3$D position and a thrust controller that computes the required thrust from this acceleration. The thrust controller is a partially open-loop controller, known as model inversion \cite{uavmodelling}. Due to its open-loop nature and dependency on the accurate values of mass and gravity, it becomes a bottleneck in aerial grasping since dynamic payloads change the mass and CoG quickly, and unmodeled external disturbances degrade its performance.
\par
Hence most works focus on improving the position feedback loop, i.e., LQR \cite{foehn2018onboard}, MPC \cite{mpc1, mpc2}, INDI \cite{indiaccurate} that requires even the motor rotational speeds in addition to the system parameters. In contrast, a recent approach~\cite{direct} replaces the open-loop thrust controller with Direct Acceleration Feedback (DA) thrust controller. However, it requires high-frequency loop rates and does not incorporate velocity control, a key requirement in aerial grasping to perform visual servoing. 
\par
While there exist a number of works for cable swinging payloads \cite{kotaru2017dynamics, tran2019adaptive,xian2020robust}; they take model-based approach, e.g. spring-mass model \cite{kotaru2017dynamics} to address their challenges which are different from ours and do not transfer to our case as they target a different problem. Moreover, they focus largely on position control but use the traditional thrust control model. On the contrary, our interest is to develop a thrust controller to address off-center aerial grasping challenges.
\par
Apart from the above, the control system's performance significantly depends on the feedback signal's accuracy and rate. Therefore, the works dealing with dynamic payloads report the control performance with very small weights by obtaining position feedback from millimeter-accurate VICON and keeping a controlled workspace. For instance, \cite{pounds2012stability} shows control of a mini helicopter for centered payload case with a tiny payload by using VICON feedback. Even DA \cite{direct} is tested with VICON. In contrast, real-world aerial grasping is far more complex since it relies only on the onboard positioning system, which is often quite slow and imprecise. Therefore, comparative evaluation of control systems is rarely available in the literature and is also a difficult task.
\par
In light of the above discussion, in this paper, we focus on aerial grasping using an off-center gripper. To the best of our knowledge, an end-to-end controller specifically targeting off-centered aerial grasping is not visible in the literature. To this end, we make the following contributions:
\par
\textit{Firstly}, our major contribution is a novel thrust controller called Thrust Microstepping via Acceleration Feedback (\texttt{TMAF}). It is highly responsive even at smaller loop rates (e.g. $30$Hz) against fast-changing system parameters and can easily tackle battery discharging and deterioration of its current deliverance that directly affects the rotor thrust (Sec.~\ref{sec:tmaf}).
\par
\textit{Secondly,} we develop an intuitive Decoupled Motion Control (\texttt{DMC}) which decouples the horizontal and vertical motion of the UAV to handle dynamic payloads (Sec.~\ref{sec:ddlmc}).  
\par
\textit{Finally,} we club \texttt{TMAF} and \texttt{DMC} into an end-to-end control system called Thrust Microstepped Decoupled Control (\OURS) for aerial grasping. \OURS{} can handle non-uniform loop rates, offers precise position and velocity control while showing highly stable flights during grasping of unknown payload (Sec.~\ref{sec:\OURS}).
\par
We benchmark \OURS{} in quite practical and adverse settings, e.g. obtaining position feedback from onboard SLAM that is slow and imprecise relative to VICON, and controlling a large quadrotor ($1.30$m$\times0.90$m$\times0.45$m) at very low altitude (below $50$cm), in narrow indoor workspace with merely $10$cm of clearance (Fig.~\ref{fig:m100}). These are quite dangerous and difficult objectives due to the risk of collision and casualty, ceiling and ground effects \cite{groundceiling}, and intense wind turbulence, however, these are common situations in aerial grasping. Noticeably, \OURS~handles them precisely, indicating the most representative accomplishment of this work (Sec.~\ref{sec:exp}). 

%

%
\section{Quadrotor System and Nomenclature}
\label{sec:sysdesc}
\begin{figure}[t]
\centering
\begin{tikzpicture}

\FPeval{\circwidth}{0.15}
\FPeval{\dotscale}{0.3}
\FPeval{\circscale}{0.5}
\colorlet{dotcolor}{white!0!green}
\colorlet{circcolor}{white!30!orange}
\colorlet{linkcolor}{black!0!orange}
\colorlet{boxdrawcolor}{white!70!orange}
\colorlet{boxfillcolor}{white!90!black}
\FPeval{\txtscale}{0.8}
\node (m100) [scale=0.85, xshift=0ex, yshift=0ex]
{
\tikz{
%
\node (m100_side) [xshift = 0ex,yshift=0ex]{\includegraphics[scale=0.07]{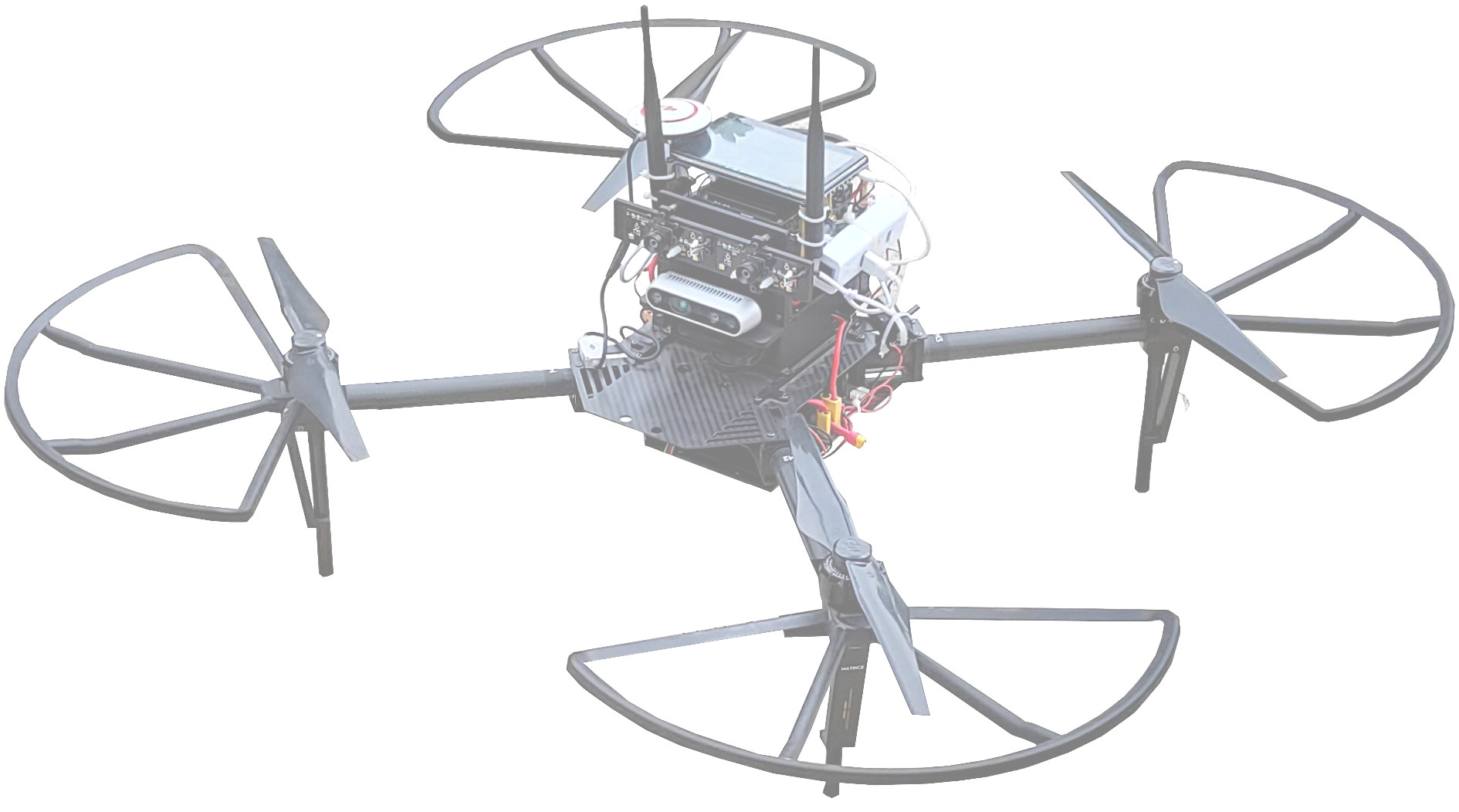}};
\node (m600_boundary) [draw=white!60!black,rounded corners=0.5mm,rectangle, align=center,minimum width=62ex,minimum height=17.5ex,xshift=0ex,yshift=0ex]{};
\node (ow) [xshift=-22.5ex,yshift=-4ex, scale=0.90]
{
\tikz{
\draw [->, red, very thick] ($(0ex,0ex)$) -- ($(3.25ex,3.25ex)$);
\draw [->, green, very thick] ($(0ex,0ex)$) -- ($(-5ex,0ex)$);
\draw [->, blue, very thick] ($(0ex,0ex)$) -- ($(-0ex,5ex)$);
\node (center) [xshift=1ex, yshift=-1.5ex]{$O_W$};
\node (x) [xshift=4.5ex, yshift=4.5ex]{$x_W$};
\node (y) [xshift=-6.5ex, yshift=0ex]{$y_W$};
\node (x) [xshift=0ex, yshift=6.5ex]{$z_W$};
}
};
\node (ob) [xshift=1ex,yshift=2ex]
{
\tikz{
\draw [->, red, very thick] ($(0ex,0ex)$) -- ($(-3.25ex,-3.25ex)$);
\draw [->, green, very thick] ($(0ex,0ex)$) -- ($(4.0ex,-1.70ex)$);
\draw [->, blue, very thick] ($(0ex,0ex)$) -- ($(-0ex,5ex)$);
\node (center) [xshift=0.5ex, yshift=-2.0ex]{$O_B$};
\node (x) [xshift=-4ex, yshift=-4ex]{$x_B$};
\node (y) [xshift=6ex, yshift=-2.5ex]{$y_B$};
\node (x) [xshift=0ex, yshift=6ex]{$z_B$};
}
};
\node (ob) [xshift=25ex,yshift=-5ex]
{
\tikz{
\draw [->, magenta, very thick] ($(0ex,0ex)$) -- ($(0ex,-5ex)$);
\node (x) [xshift=1.2ex, yshift=-2.5ex]{$\bm g$};
}
};
}
};
%
\end{tikzpicture}
\vspace{-1.5ex}
\caption{Quadrotor reference frames.}
\label{fig:m100wogripper}
\vspace{-3.0ex}
\end{figure}
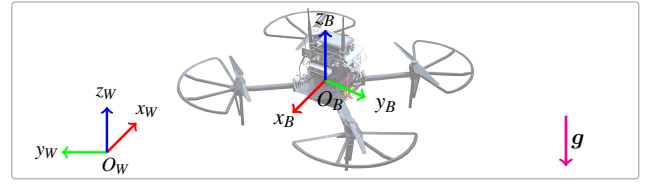

Our quadrotor platform is shown in Fig.~\ref{fig:m100wogripper}. We define a \textit{world frame} $\mathcal{F}_W$ having axes $\{\bm x_W, \bm y_W, \bm z_W \}$ that is always fixed at the take-off location. The gravity vector  $\bm g \in \mathbb{R}^3$ is aligned opposite to the $\bm z_W$ axis. While the quadrotor configuration is described via a \textit{body frame} $\mathcal{F}_B$, which coincides with the Center-of-Gravity (CoG) of the quadrotor, and has axes $\{\bm x_B, \bm y_B, \bm z_B \}$. Both the frames $\mathcal{F}_W$ and $\mathcal{F}_B$ are FLU or \textit{\textcolor{red}{Front}-\textcolor{green}{Left}-\textcolor{blue}{Up}}, denoting their $\{\bm x, \bm y, \bm z \}$ axes. 
\par
The position and orientation of $\mathcal{F}_B$ relative to $\mathcal{F}_W$ is given by a vector $\bm p_B=[x, y, z]^T \in \mathbb{R}^3$ and a rotation matrix $\bm R_B \in SO3$. The quadrotor's velocity, acceleration, and angular velocity, are represented as $\bm v_B=[v_x, v_y, v_z]^T \in \mathbb{R}^3$, $\bm a_B=[a_x, a_y, a_z]^T \in \mathbb{R}^3$, $\bm \omega_B=[\omega_x, \omega_y, \omega_z]^T \in \mathbb{R}^3$, expressed in $\mathcal{F}_W$, except that $\omega_B$ is in $\mathcal{F}_B$. \textit{Note:} The \textit{boldface} denotes a vector.
\par
The quadrotor motion is governed by the \textit{thrust intensity} $ f_B \in \mathbb{R}$, and \textit{torques} $\bm \tau_B =[\tau_x, \tau_y, \tau_z]^T \in \mathbb{R}^3$. $f_B$ is applied parallel to $\bm z_B$ whereas $\bm \tau_B$ operates in $\mathcal{F}_B$ which controls the roll-pitch-yaw of the quadrotor in $\mathcal{F}_B$. Practically, controlling the quadrotor's position/velocity and heading ($\psi$) is of great interest, which depends on $\{f_B, \phi, \theta, \omega_z\}$. Here, $\phi$ and $\theta$ are defined in $\mathcal{F}_B$ whereas $\psi$ in $\mathcal{F}_W$. The position controller generates $\{f_B^\star, \phi^\star, \theta^\star\}$ commands based on the desired state, i.e. desired position/velocity and heading ($\psi^\star$). The `$\star$' superscript denotes the desired value of any variable. 
%
\par
These commands are sent to the low-level attitude controller, which essentially controls the angular velocities of the platform to generate $\bm \tau^\star_B$ for achieving the desired $\{\phi^\star, \theta^\star, \psi^\star\}$. Finally, based on the desired thrust and the torques, motor speed allocation is performed \cite{uavmodelling}, which in turn is realized via Electronic-Speed-Controller (ESC).
\par
In the quadrotor's position control process, the thrust controller or \textit{Model Inversion} (Eq.~\ref{eq:classicalthrust}) plays a crucial role.
\begin{equation}
\footnotesize
\bm f^\star_B = m (\bm a^\star_B + \bm g) -  \bm f_e
\label{eq:classicalthrust}
\end{equation}
where, $\bm f_e \in \mathbb{R}^{3}$ is an external disturbance (wind draft, sudden jerk, imperfect modelling), $m$ is the UAV mass, and $\bm a^\star_B$ is  the desired acceleration produced by the position feedback loop. Since motor thrust is applicable only along $\bm z_B$, a desired orientation of the quadrotor is also required to attain $\bm f^\star_B$. It is computed via geometric tracking controller \cite{geometrictracking} as follows:
\begin{equation}
\footnotesize
\bm z^\star_B = \bm f^\star_B / \norm{\bm f^\star_B}
\label{eq:forcedir}
\end{equation}
By using the Eq.~\ref{eq:classicalthrust} and \ref{eq:forcedir}, the desired thrust intensity $f^\star_B = \bm z^T_B \bm f^\star_B$, and $\{\phi^\star, \theta^\star\}$ can be computed \cite{geometrictracking}, whereas the desired heading $\psi^\star$ is obtained as input from the user.
\par
As visible in Eq.~\ref{eq:classicalthrust} that the model inversion depends on the mass, gravity and disturbance modeling to estimate $f^\star_B$ accurately, it may lead to incorrect thrust estimation and, thus, undesired response in dynamic payload conditions of aerial grasping. Hence our problem of interest is to develop a mass and gravity-independent thrust controller which can also handle low-frequency loop rates and non-uniform execution. 
%

\vspace{-0ex}
\section{Thrust Microstepping}
\label{sec:tmaf}
\begin{figure}[t]
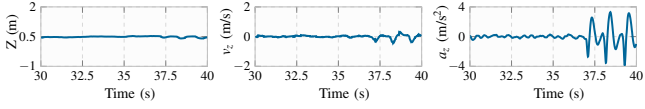

\centering

\begin{tikzpicture}

\FPeval{\xshfta}{0}
\FPeval{\xshftb}{0+18}
\FPeval{\xshftc}{0+36}
\FPeval{\xshftd}{0}

\FPeval{\yshfta}{0-0}
\FPeval{\yshftb}{0-0}
\FPeval{\yshftc}{0-0}
\FPeval{\yshftd}{0-0}

\FPeval{\pltw}{38}
\FPeval{\plth}{20}

\FPeval{\scal}{0.5}

\FPeval{\mrksize}{0.3}

\colorlet{desclr}{white!0!magenta}
\colorlet{currclr}{blue!60!green}

\colorlet{gridclr}{white!85!black}
\colorlet{dlegendclr}{white!80!black}
\colorlet{axisclr}{white!75!black}
\colorlet{axisbgclr}{white!99!black}

\colorlet{dlegendclr}{white!80!black}
\colorlet{legendclr}{white!100!black}

\FPeval{\lscale}{0.6}
\FPeval{\limscale}{0.5}
\FPeval{\lxshft}{5.5}
\FPeval{\lyshft}{9.0}

\FPeval{\titlescale}{0.7}
\FPeval{\titlexshift}{21.0}
\FPeval{\titleyshift}{13.7}

\colorlet{gpudclr}{white!80!black}
\colorlet{gpuclr}{white!90!black}
\colorlet{gputxtclr}{white!0!black}

\FPeval{\labelscale}{1.1}
\FPeval{\ticklabelscale}{1.0}

\FPeval{\xlabelxshift}{0+15}
\FPeval{\xlabelyshift}{0+0.5}
\FPeval{\ylabelxshift}{0+3.0}
\FPeval{\ylabelyshift}{0+5.8}

\FPeval{\linew}{1.5}
\FPeval{\dashon}{5.0}
\FPeval{\dashoff}{3.0}

\FPeval{\xshfttakeoff}{0-16}
\FPeval{\xshfthover}{0}
\FPeval{\xshftland}{16}
\node (a) [xshift = \xshfta ex, yshift=\yshfta ex, scale=\scal]{
\tikz{
\pgfplotsset{width=\pltw ex, height=\plth ex}
\begin{axis}[
   axis background style={fill=axisbgclr},
    title={},
    xlabel={Time (s)},
    ylabel={ Z (m)},
    xmin=30, xmax=40,
    ymin=-1, ymax=2,
    xtick={30, 32.5, 35, 37.5, 40},
    ytick={-1, 0.5,  2},
    xticklabels={$30$, $32.5$, $35$, $37.5$, $40$},
    yticklabels={$-1$, $0.5$, $2$},
     axis line style={axisclr},
    legend image post style={scale =\limscale},
    legend style={at={(\lxshft ex,\lyshft ex)},anchor=south, legend columns = 1, draw = {dlegendclr}, fill={legendclr}, nodes={scale=\lscale}},
    ymajorgrids=true, 
    xmajorgrids=true,
    grid style={dashed, gridclr},
    major tick length=1ex,
    x label style={at={(\xlabelxshift ex, \xlabelyshift ex)},scale=\labelscale},
    y label style={at={(\ylabelxshift ex, \ylabelyshift ex)},scale=\labelscale},
    xticklabel style={scale=\ticklabelscale},
    yticklabel style={scale=\ticklabelscale},
]
\input{exp_plots/z_disturbance/z}
\end{axis}
%
}};
\node (b) [xshift = \xshftb ex, yshift=\yshftb ex, scale=\scal]{
\tikz{
\pgfplotsset{width=\pltw ex, height=\plth ex}
\begin{axis}[
   axis background style={fill=axisbgclr},
    title={},
    xlabel={Time (s)},
    ylabel={ $v_z$ (m/s)},
    xmin=30, xmax=40,
    ymin=-2, ymax=2,
    xtick={30, 32.5, 35, 37.5, 40},
    ytick={-2, 0, 2},
    xticklabels={$30$, $32.5$, $35$, $37.5$, $40$},
    yticklabels={$-2$, $0$, $2$},
     axis line style={axisclr},
    legend image post style={scale =\limscale},
    legend style={at={(\lxshft ex,\lyshft ex)},anchor=south, legend columns = 1, draw = {dlegendclr}, fill={legendclr}, nodes={scale=\lscale}},
    ymajorgrids=true, 
    xmajorgrids=true,
    grid style={dashed, gridclr},
    major tick length=1ex,
    x label style={at={(\xlabelxshift ex, \xlabelyshift ex)},scale=\labelscale},
    y label style={at={(\ylabelxshift ex, \ylabelyshift ex)},scale=\labelscale},
    xticklabel style={scale=\ticklabelscale},
    yticklabel style={scale=\ticklabelscale},
]
\input{exp_plots/z_disturbance/vz}
\end{axis}
%
}};
\node (c) [xshift = \xshftc ex, yshift=\yshftc ex, scale=\scal]{
\tikz{
\pgfplotsset{width=\pltw ex, height=\plth ex}
\begin{axis}[
   axis background style={fill=axisbgclr},
    title={},
    xlabel={Time (s)},
    ylabel={ $a_{z}$ (m/s$^2$)},
    xmin=30, xmax=40,
    ymin=-4, ymax=4,
    xtick={30, 32.5, 35, 37.5, 40},
    ytick={-4, 0, 4},
    xticklabels={$30$, $32.5$, $35$, $37.5$, $40$},
    yticklabels={$-4$, $0$, $4$},
     axis line style={axisclr},
    legend image post style={scale =\limscale},
    legend style={at={(\lxshft ex,\lyshft ex)},anchor=south, legend columns = 1, draw = {dlegendclr}, fill={legendclr}, nodes={scale=\lscale}},
    ymajorgrids=true, 
    xmajorgrids=true,
    grid style={dashed, gridclr},
    major tick length=1ex,
    x label style={at={(\xlabelxshift ex, \xlabelyshift ex)},scale=\labelscale},
    y label style={at={(\ylabelxshift ex, \ylabelyshift ex)},scale=\labelscale},
    xticklabel style={scale=\ticklabelscale},
    yticklabel style={scale=\ticklabelscale},
]
\input{exp_plots/z_disturbance/az}
%
%
    %
\end{axis}
%
}};

\end{tikzpicture}
\vspace{-1.5ex}
\caption{Effect of a sudden force of $12$N in $\bm z_W$ onto the $z, v_z, a_z$ of the quadrotor. Noticeably, the acceleration signal captures a detailed profile of this event, while position and velocity signals do not.}
\label{fig:zdisturbancesample}
\vspace{-2.0ex}
\end{figure}
\begin{figure}[t]
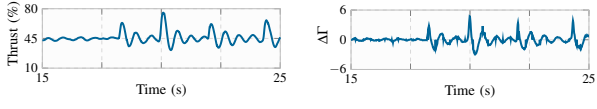

\centering

\begin{tikzpicture}

\FPeval{\xshfta}{0}
\FPeval{\xshftb}{0+26}
\FPeval{\xshftc}{0+36}
\FPeval{\xshftd}{0}

\FPeval{\yshfta}{0-0}
\FPeval{\yshftb}{0-0.25}
\FPeval{\yshftc}{0-0}
\FPeval{\yshftd}{0-0}

\FPeval{\pltw}{50}
\FPeval{\plth}{20}

\FPeval{\scal}{0.5}

\FPeval{\mrksize}{0.3}

\colorlet{desclr}{white!0!magenta}
\colorlet{currclr}{blue!60!green}

\colorlet{gridclr}{white!85!black}
\colorlet{dlegendclr}{white!80!black}
\colorlet{axisclr}{white!75!black}
\colorlet{axisbgclr}{white!99!black}

\colorlet{dlegendclr}{white!80!black}
\colorlet{legendclr}{white!100!black}

\FPeval{\lscale}{0.6}
\FPeval{\limscale}{0.5}
\FPeval{\lxshft}{5.5}
\FPeval{\lyshft}{9.0}

\FPeval{\titlescale}{0.7}
\FPeval{\titlexshift}{21.0}
\FPeval{\titleyshift}{13.7}

\colorlet{gpudclr}{white!80!black}
\colorlet{gpuclr}{white!90!black}
\colorlet{gputxtclr}{white!0!black}

\FPeval{\labelscale}{1.1}
\FPeval{\ticklabelscale}{1.0}

\FPeval{\xlabelxshift}{0+20}
\FPeval{\xlabelyshift}{0+1.5}
\FPeval{\ylabelxshift}{0+3.0}
\FPeval{\ylabelyshift}{0+5.8}

\FPeval{\linew}{1.5}
\FPeval{\dashon}{5.0}
\FPeval{\dashoff}{3.0}

\FPeval{\xshfttakeoff}{0-16}
\FPeval{\xshfthover}{0}
\FPeval{\xshftland}{16}
\node (a) [xshift = \xshfta ex, yshift=\yshfta ex, scale=\scal]{
\tikz{
\pgfplotsset{width=\pltw ex, height=\plth ex}
\begin{axis}[
   axis background style={fill=axisbgclr},
    title={},
    xlabel={Time (s)},
    ylabel={ Thrust ($\%$)},
    xmin=0, xmax=1.0,
    ymin=10, ymax=80,
    xtick={0, 0.25, 0.5, 0.75, 1.0},
    ytick={10, 45,  80},
    xticklabels={$15$, $ $, $ $, $ $, $25$},
    yticklabels={$10$, $45$, $80$},
     axis line style={axisclr},
    legend image post style={scale =\limscale},
    legend style={at={(\lxshft ex,\lyshft ex)},anchor=south, legend columns = 1, draw = {dlegendclr}, fill={legendclr}, nodes={scale=\lscale}},
    ymajorgrids=true, 
    xmajorgrids=true,
    grid style={dashed, gridclr},
    major tick length=1ex,
    x label style={at={(\xlabelxshift ex, \xlabelyshift ex)},scale=\labelscale},
    y label style={at={(\ylabelxshift ex, \ylabelyshift ex)},scale=\labelscale},
    xticklabel style={scale=\ticklabelscale},
    yticklabel style={scale=\ticklabelscale},
]
\input{exp_plots/microstep/thrust}
\end{axis}
%
}};
\node (b) [xshift = \xshftb ex, yshift=\yshftb ex, scale=\scal]{
\tikz{
\pgfplotsset{width=\pltw ex, height=\plth ex}
\begin{axis}[
   axis background style={fill=axisbgclr},
    title={},
    xlabel={Time (s)},
    ylabel={$\Delta \Gamma$},
    xmin=0, xmax=1.0,
    ymin=-6, ymax=6,
    xtick={0, 0.25, 0.5, 0.75, 1.0},
    ytick={-6, 0,  6},
    xticklabels={$15$, $ $, $ $, $ $, $25$},
    yticklabels={$-6$, $0$, $6$},
     axis line style={axisclr},
    legend image post style={scale =\limscale},
    legend style={at={(\lxshft ex,\lyshft ex)},anchor=south, legend columns = 1, draw = {dlegendclr}, fill={legendclr}, nodes={scale=\lscale}},
    ymajorgrids=true, 
    xmajorgrids=true,
    grid style={dashed, gridclr},
    major tick length=1ex,
    x label style={at={(\xlabelxshift ex, \xlabelyshift ex)},scale=\labelscale},
    y label style={at={(\ylabelxshift ex, \ylabelyshift ex)},scale=\labelscale},
    xticklabel style={scale=\ticklabelscale},
    yticklabel style={scale=\ticklabelscale},
]
\input{exp_plots/microstep/microsteps}
\end{axis}
%
}};

\end{tikzpicture}
\vspace{-0.8ex}
\caption{Thrust intensity ($f^\star_B$) and microsteps ($\Delta \Gamma$) output of \texttt{TMAF}.}
\label{fig:microstep}
\vspace{-3.0ex}
\end{figure}
We propose Thrust Microstepping via Acceleration Feedback (\texttt{TMAF}) to estimate the desired thrust. The microstepping is inspired by the microstepping technique for fine-grained control of the stepper motors. Whereas the use of acceleration feedback is based on our observations that during payload attach, detach, or disturbance events, the acceleration signal changes rapidly, even at low-frequency sampling and small variations in the position or velocity signal (Fig.~\ref{fig:zdisturbancesample}). We divide \texttt{TMAF} into two steps, as described below.
\subsubsection{Microstepper Controller}
In this step, we generate a microstep $\bm \Delta \bm \Gamma$ based on the position feedback loop's output $\bm a^\star_B$. For that, we define the following control law:
 \begin{equation}
 \footnotesize
 \bm \Delta \bm \Gamma =  \alpha \bm e_{\bm a} + \beta \dot{\bm e}_{\bm a}
\label{eq:tmafmisctrl}
 \end{equation}
where, $\bm e_{\bm a} = \bm a^\star_B - \bm a_B$ is the acceleration error, and $\bm \alpha, \bm \beta$ are the controllers gains. Notice that, this controller essentially tracks $\bm a^\star_B$ and fully captures the variations in the acceleration signal. When $\bm a_B$ undergoes large changes, $\bm e_{\bm a}$ becomes large. Thus, $\bm \Delta \bm \Gamma$ changes accordingly to handle the changes.
\subsubsection{Microstepped Thrust Estimation}
This step computes the total instantaneous thrust $\bm \Gamma_t$ or $\bm f^\star_B$ by accumulating past $\bm \Delta \Gamma$'s in a recursively i.e $\bm \Gamma_t = \bm \Gamma_{t-1} + \bm \Delta \Gamma_t$, where, $\bm \Gamma_{t-1} = \sum^{t-1}_{i=0} \bm \Delta \Gamma_{i}$. Now, rewriting the previous expression as follows:
\begin{equation}
\footnotesize
\bm f^\star_B \equiv \bm \Gamma_t = \bm \Gamma_{0} + \sum^{t}_{i=1} \bm \Delta \Gamma_{i} \implies \bm f^\star_B = \bm f^\star_{B_{t-1}}  + \bm \Delta \Gamma_{t}
\label{eq:thrusttmaf}
\end{equation}
where, $\bm \Gamma_0$$=$$0$, and $t$$=$$0$ indicates flight commencement. For the convergence proof, please refer to Appendix~\ref{app:proof}.
\par
The accumulation in Eq.~\ref{eq:thrusttmaf} should not be confused with the integrator in the PID control because `$dt$' is absent here. This enables \texttt{TMAF} to exploit rapid acceleration changes to counter sudden disturbances instantly (Fig.~\ref{fig:microstep}, Fig.~\ref{fig:tmafvsda_disturb}). Now $\bm f^\star_B = \{f^\star_{x},f^\star_{y},f^\star_{z}\}$ can be used to obtain $\{f_B^\star, \phi^\star, \theta^\star\}$ via Eq.~\ref{eq:forcedir}. 

\begin{figure}[t]
\centering

\begin{tikzpicture}

\FPeval{\inscal}{0.9}

\FPeval{\cscal}{1.0}

\node (outer) [scale=0.55]
{
\tikz{
\node (outer) [draw=white!50!gray,rounded corners=0.2ex, minimum width=98.5ex, minimum height=26.5ex, xshift=5.5ex, yshift=-9ex]{};
\node (a) [draw= none, rounded corners=0.2ex,xshift=-26ex, yshift=-9.5ex, scale=\inscal] {$\bm a^\star_B$};
\node (psi) [draw= none, rounded corners=0.2ex,xshift=25.5ex, yshift=0ex, scale=\inscal] {$\psi^\star$};
\node (mi) [draw= none, yshift=0ex, scale=\cscal]
{
\tikz{
\node (outermi) [draw=cyan, fill=white!80!cyan, dashed, dash pattern=on 5pt off 2pt, line width=0.1ex, rounded corners=0.2ex, minimum width=28ex, minimum height=7ex, xshift=0ex, yshift=1.5ex,]{};
\node (textmi)[xshift=0ex, yshift=3.8ex]{\texttt{Model Inversion}};
\node (eqmi) [draw= gray,fill=white, rounded corners=0.2ex, xshift=1ex, scale=\inscal] {$\bm f^\star=m(\bm a^\star_B + \bm g) + \bm f_e$};
}
};
\node (da) [xshift=0, yshift=-8ex, scale=\cscal]
{
\tikz{
\node (outerda) [draw=blue, fill=white!80!blue, dashed, dash pattern=on 5pt off 2pt, line width=0.1ex, rounded corners=0.2ex, minimum width=28ex, minimum height=7ex, xshift=18.8ex, yshift=1.5ex,]{};
\node (textda)[xshift=18.8ex, yshift=3.8ex]{\texttt{DA}};
\node (sumda) [draw= none, rounded corners=0.2ex, xshift=10ex, scale=\inscal] {\tikz{\node[draw=gray,fill=white, circle,scale=1.2]{};\node[draw=none, scale=1.3]{$+$};}};
\node (ahat) [draw=none, xshift=10ex, yshift=3.5ex, scale=\inscal]{$\bm g$};
\node (eqda) [draw= gray,fill=white, rounded corners=0.2ex, xshift=22ex, scale=\inscal] {$\bm f^\star = \bm \mu \bm a^\star_B + \bm \lambda \int \bm e_{\bm a}$};
%
\draw [->] ($(ahat.south)+(0.0ex, 0.4ex)$) -- ($(sumda.north)-(0.0ex, 0.8ex)$);
\draw [->] ($(sumda.east)+(-0.6ex, 0.0ex)$) -- ($(eqda.west)-(0.0ex, 0.0ex)$);
}
};
\node (tmaf) [xshift=0, yshift=-17ex, scale=\cscal]
{
\tikz{
\node (outertmaf) [draw=magenta, fill=white!80!gray, dashed, dash pattern=on 5pt off 2pt, line width=0.1ex, rounded corners=0.2ex, minimum width=28ex, minimum height=9ex, xshift=18.8ex, yshift=9.5ex,]{};
\node (texttmaf)[xshift=18.8ex, yshift=12.8ex, scale=1.0]{\texttt{TMAF}};
\node (eqtmaf) [draw= gray, fill=white, rounded corners=0.2ex, xshift=13ex,yshift=9.5ex,  scale=\inscal] {$\bm \alpha \bm e_{\bm a} + \bm \beta \dot{\bm e_{\bm a}}$};
\node (sumtmaf) [draw= none, rounded corners=0.2ex, xshift=23ex, yshift=9.5ex, scale=\inscal] {\tikz{\node[draw=gray,fill=white, circle,scale=1.2]{};\node[draw=none, scale=1.3]{$+$};}};
\node (tt1) [draw= none, rounded corners=0.2ex, xshift=23ex, yshift=6ex, scale=\inscal] {$\bm u_{t-1}$};
\node (tt) [draw= none, rounded corners=0.2ex, xshift=28ex, yshift=11.5ex, scale=\inscal] {$\bm u_{t} \equiv \bm f^\star$};
\draw [->] ($(eqtmaf.east)+(0.0ex, 0.0ex)$) -- ($(sumtmaf.west)+(0.8ex, 0.0ex)$);
\draw [->] ($(tt1.north)-(0.0ex, 0.7ex)$) -- ($(sumtmaf.south)+(0.0ex, 0.8ex)$);
}
};
\node (gt) [draw=black, rounded corners=0.2ex, xshift=25.5ex, yshift=-9.5ex, scale=0.8]{\shortstack{Geometric \\ Tracking}};
\node (output)[xshift=37.5ex, yshift=-9.5ex, scale=\inscal]
{
\tikz{
\node (fmag) [draw= black,xshift=0ex, yshift=0ex, scale=1.0] {\shortstack{$\phi^\star$ \\ $\theta^\star$ \\ $\psi^\star$ \\ $f^\star$}};
\node  [fill=white,yshift=6ex, minimum width=2.0ex]{};
\node  [fill=white,yshift=-6.0ex, minimum width=2.0ex]{};
}};
\colorlet{arrowclr}{green!40!blue}
\draw [->, arrowclr, dashed, dash pattern=on 4pt off 1.7pt] ($(a.east)+(0ex, 0ex)$) .. controls ($(a.east)+(8ex, 1ex)$) and ($(mi.west)-(2.7ex,1.0ex)$) .. ($(mi.west)+(6.7ex,-1.7ex)$);
\draw [->, arrowclr, dashed, dash pattern=on 4pt off 1.7pt] ($(a.east)+(0ex, 0ex)$) .. controls ($(a.east)+(8ex, 0ex)$) and ($(da.west)+(1ex,-1.5ex)$) .. ($(da.west)+(5ex,-1.5ex)$);
\draw [->, arrowclr, dashed, dash pattern=on 4pt off 1.7pt] ($(a.east)+(0ex, 0ex)$) .. controls ($(a.east)+(8ex, -1ex)$) and ($(tmaf.west)-(5ex, 0ex)$) .. ($(tmaf.west)+(3.9ex, 0.0ex)$);
\draw [, arrowclr, dashed, dash pattern=on 4pt off 1.7pt] ($(mi.east)-(4.5ex, 1.7ex)$) .. controls ($(mi.east)-(0.5ex, 1.2ex)$) and ($(gt.west)-(3ex, 0ex)$) .. ($(gt.west)$);
\draw [->, arrowclr, dashed, dash pattern=on 4pt off 1.7pt] ($(da.east)-(3.0ex, 1.5ex)$) .. controls ($(da.east)-(0.0ex, 1.5ex)$) and ($(gt.west)-(5ex,0ex)$) .. ($(gt.west)$);
\draw [, arrowclr, dashed, dash pattern=on 4pt off 1.7pt] ($(tmaf.east)-(9.5ex, 0ex)$) .. controls ($(tmaf.east)+(1ex, -1.1ex)$) and ($(gt.west)-(2ex, 0ex)$) .. ($(gt.west)$);
\draw [->, ] ($(psi.south)+(0ex, 0ex)$) -- ($(gt.north)$);
\draw [->, ] ($(gt.east)+(0ex, 0ex)$) -- ($(output.west)$);
}};

\end{tikzpicture}
\vspace{-0.5ex}
\caption{Thrust Microstepping via Acceleration Feedback (\texttt{TMAF}) \textit{vs} Direct Acceleration (DA) Feedback \cite{direct} and Model Inversion \cite{uavmodelling}.}
\label{fig:tmafvsdaarch}
\vspace{-2.8ex}
\end{figure}
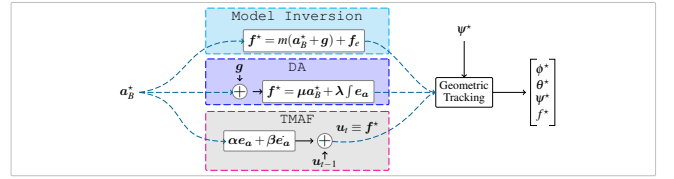

\subsection{Structural Analysis of \texttt{TMAF} and State-of-the-art}
\label{sec:da}
Model inversion (MI) (Eq.~\ref{eq:classicalthrust}) and DA \cite{direct} are the most suitable approaches to compare with \texttt{TMAF}. For better analysis, we write formulations of the three controllers below:
\par
{
\footnotesize
\begin{align}
\text{MI} \equiv \bm f^\star_B &= m (\bm a^\star_B + \bm g) -  \bm f_e\label{eq:forceclassic} \\
\text{DA} \equiv \bm f^\star_B &= \bm \mu \bm a^\star_B + \bm \lambda \int \bm e_{\bm a} dt \label{eq:forceda} \\
\texttt{TMAF} \equiv \bm f^\star_B &= \alpha \bm e_{\bm a} + \beta \dot{\bm e}_{\bm a} + \bm f^{\star}_{B_{t-1}} \label{eq:forcetmaf}
\end{align}
}
where $\bm \mu, \bm \lambda, \bm \alpha, \bm \beta$ are the tunable parameters, $ \bm e_{\bm a} = \bm a^\star_B - \bm a_B$, and $\bm f^\star_B$ is the desired thrust vector to attain $\bm a^\star_B$. If comparing Eq.~\ref{eq:forceda} with Eq.~\ref{eq:forceclassic}, it can be noticed that $\bm \mu$ absorbs the mass and gravity, while $\bm \lambda$ handles the uncertainties and external disturbances \cite{direct}. Whereas our \texttt{TMAF} is substantially different from the model inversion and DA, as shown in Fig.~\ref{fig:tmafvsdaarch}.
\subsubsection{Absence of Mass and Gravity}
Similar to model inversion, DA also needs gravity compensation that is added to $\bm a^\star_B$ (Fig.~\ref{fig:tmafvsdaarch}). Whereas \texttt{TMAF} is completely unaware of the system mass and gravity. This makes \texttt{TMAF} free of uncertainty in these parameters, allowing \texttt{TMAF} to offer precise control.
\subsubsection{Absence of Error Integral}
DA (Eq.~\ref{eq:forceda}) is a Proportional-Integral (PI) controller. At low-frequency loop rates, the integral term turns DA slower due to the low pass filtering effects and also needs integral windup. The low-frequency loop rates also prevent using larger values of $\bm \lambda$ in DA to achieve faster response, because, it causes system oscillations due to the enhanced integrator's dominance \cite{direct}. Therefore, DA should be run at high loop rates ($>250$Hz). However, in this case, the accelerometer noise jumps-in, which adversely affects the DA performance, thus mandating minimum-lag filtering of the acceleration measurements \cite{direct}.
\par
In contrast, \texttt{TMAF} only has difference terms (Eq.~\ref{eq:tmafmisctrl}) which alleviates runtime bias estimation, and the absence of an integral term turns \texttt{TMAF} highly responsive while avoiding the integral windup. Therefore, \texttt{TMAF} can seamlessly work even at a low frequency which naturally acts as a low pass filter \cite{direct} on the accelerometer signal (Sec.~\ref{sec:exp_loop_rates}). Thus unlike DA, our \texttt{TMAF} does not need acceleration filtering, saving overhead when using \texttt{TMAF} in microcontrollers.
\par
Moreover, the Eq.~\ref{eq:tmafmisctrl} consists of an acceleration differential term (Jerk), making \texttt{TMAF} instantly reactive to sudden disturbances. This term may have high-frequency noise, but the low-frequency operation implicitly handles that.

\subsubsection{Rapid Acceleration Changes}
\label{sec:tmaf_rapid_changes}
The absence of integral term allows \texttt{TMAF} to exploit rapid changes in the acceleration. This is why \texttt{TMAF} shows stiffness against sudden disturbances (See Sec.~\ref{sec:distrub_tmaf_da} and video). In contrast, MI (Eq.~\ref{eq:classicalthrust}) can not harness them due to its open-loop nature, and DA, by nature, suppresses them due to its integral action over $e_a$ (Eq.~\ref{eq:forceda}).
\subsection{\texttt{TMAF} in Aerial Manipulation}
In aerial grasping, the dynamic payloads alter the system mass, and the battery discharging and ageing reduce the current output. Since the system parameters are still the same, the above events reduce rotor thrust via Eq.~\ref{eq:classicalthrust}. To compensate, the thrust should be corrected accordingly. However, in model inversion \cite{uavmodelling}, it is done manually via hit-and-trial methods and is kept fixed during the flight ($50\%$ in \cite{pixhawk}). While in DA, the value of $\bm \mu$ needs to be adjusted.
\par
In contrast, \texttt{TMAF} captures these events via accelerometer readings since inaccurate thrust would cause undesired motion, and adjusts the thrust via microstepping without requiring manual tuning (Fig.~\ref{fig:hoverdynamicpayload} and See video). 
%

%
 %
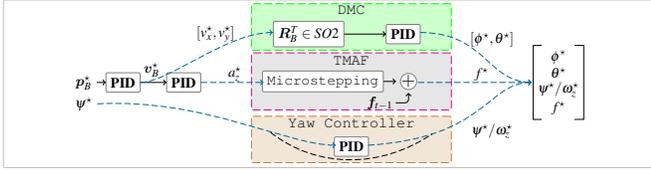
\begin{figure}[t]
\centering
\begin{tikzpicture}

\FPeval{\inscal}{0.9}

\FPeval{\cscal}{1.0}

\node (adpd) [draw=none,xshift=0.0ex,yshift=0ex, scale=0.60]
{
\tikz{
\node (outer) [draw=white!50!gray,rounded corners=0.2ex, minimum width=91.5ex, minimum height=23.5ex, xshift=5.5ex, yshift=-8.2ex]{};
\node (p) [draw= none, rounded corners=0.2ex,xshift=-29ex, yshift=-8ex, scale=\inscal] {$\bm p^\star_B$};
\node (psi) [draw= none, rounded corners=0.2ex,xshift=-29.0ex, yshift=-11.0ex, scale=\inscal] {$\psi^\star$};
\node (pidp) [draw= gray, fill=white, fill=white, rounded corners=0.2ex,xshift=-23.5ex,yshift=-8ex, scale=\inscal] {$\textbf{PID}$};
\node (pidv) [draw= gray,  fill=white,rounded corners=0.2ex, xshift=-15ex,yshift=-8ex,  scale=\inscal] {$\textbf{PID}$};
\node (ddlmc) [xshift=8.5ex,yshift=0ex, scale=\cscal]
{
\tikz{
\node (outerddlmc) [draw=green, fill=white!80!green, dashed, dash pattern=on 5pt off 2pt, line width=0.1ex, rounded corners=0.2ex, minimum width=28ex, minimum height=6.5ex, xshift=0.0ex, yshift=0.0ex,]{};
\node ()[xshift=0.0ex, yshift=2.4ex, scale=0.95]{\texttt{DMC}};
%
\node (rso2) [draw= gray, fill=white, rounded corners=0.2ex, xshift=-6 ex, yshift = -1.0ex, scale=\inscal] {$\bm R^T_B \in SO2$};
\node (pidvxy) [draw= gray, fill=white, rounded corners=0.2ex, xshift=7.25 ex, yshift = -1.0ex, scale=\inscal] {$\textbf{PID}$};
\draw [->] ($(rso2.east)+(0.0ex, 0.0ex)$) -- ($(pidvxy.west)+(0.0ex, 0.0ex)$);
}
};
\node (tmaf) [xshift=8.5ex, yshift=-8.0ex, scale=\cscal]
{
\tikz{
\node (outertmaf) [draw=magenta, fill=white!80!gray, dashed, dash pattern=on 5pt off 2pt, line width=0.1ex, rounded corners=0.2ex, minimum width=28ex, minimum height=8.0ex, xshift=0.0ex, yshift=0.0ex,]{};
\node ()[xshift=0ex, yshift=3.0ex, scale=0.95]{\texttt{TMAF}};
\node (eqtmaf) [draw= gray, fill=white, rounded corners=0.2ex, xshift=-4ex,yshift=0ex,  scale=\inscal] {\texttt{Microstepping}};
\node (sumtmaf) [draw= none, rounded corners=0.2ex, xshift=8ex, yshift=0ex, scale=\inscal] {\tikz{\node[draw=gray,fill=white, circle,scale=1.2]{};\node[draw=none, scale=1.3]{$+$};}};
\node (tt1) [draw= none, rounded corners=0.2ex, xshift=4ex, yshift=-3ex, scale=\inscal] {$\bm f_{t-1}$};
\draw [->] ($(eqtmaf.east)+(0.0ex, 0.0ex)$) -- ($(sumtmaf.west)+(0.8ex, 0.0ex)$);
\draw [->, rounded corners=0.5ex] ($(tt1.east)-(0.3ex, 0.0ex)$) -| ($(sumtmaf.south)+(0.0ex, 0.8ex)$);
}
};
\node (yaw) [xshift=8.5ex, yshift=-15.8ex, scale=\cscal]
{
\tikz{
\node (outeryaw) [draw=brown, fill=white!80!brown, dashed, dash pattern=on 5pt off 2pt, line width=0.1ex, rounded corners=0.2ex, minimum width=28ex, minimum height=6.5ex, xshift=0.0ex, yshift=0.0ex,]{};
\node ()[xshift=0.0ex, yshift=2.3ex, scale=0.95]{\texttt{Yaw Controller}};
\node (pidyaw) [draw= gray, fill=white, rounded corners=0.2ex, xshift=0 ex, yshift = -1.0ex, scale=\inscal] {$\textbf{PID}$};
}
};
\node (output)[xshift=37.5ex, yshift=-8ex, scale=\inscal]
{
\tikz{
\node (fmag) [draw= black,xshift=0ex, yshift=0ex, scale=1.0] {\shortstack{$\phi^\star$ \\ $\theta^\star$ \\ $\psi^\star/\omega^\star_{z}$ \\ $f^\star$}};
\node (a) [fill=white,yshift=6ex, minimum width=5.0ex]{};
\node (b) [fill=white,yshift=-6.0ex, minimum width=5.0ex]{};
}};
\draw [->] (p) -- (pidp);
\draw [->] (pidp) -- (pidv) node ()[xshift=-4.5ex, yshift=1.8ex,scale=\inscal]{$\bm v^\star_B$};
\colorlet{arrowclr}{green!40!blue}
\draw [->, arrowclr, dashed, dash pattern=on 4pt off 1.7pt] ($(pidp.east)+(0ex, 0ex)$) .. controls ($(pidp.east)+(4ex, -0.0ex)$) and ($(ddlmc.west)-(2.7ex,1.0ex)$) .. ($(ddlmc.west)+(3.8ex,-1.0ex)$) node [black, xshift=-8ex, yshift=-0.2ex, scale=\inscal]{$[v^\star_x, v^\star_y]$};
\draw [->, arrowclr, dashed, dash pattern=on 4pt off 1.7pt] ($(pidv.east)+(0ex, 0ex)$) .. controls ($(pidv.east)+(4ex, -0.0ex)$) and ($(tmaf.west)-(2.7ex,0.0ex)$) .. ($(tmaf.west)+(2.5ex,0.0ex)$) node [black, xshift=-4ex, yshift=1.2ex, scale=\inscal]{$a^\star_z$};
\draw [->, arrowclr, dashed, dash pattern=on 4pt off 1.7pt] ($(psi.east)+(0ex, 0ex)$) .. controls ($(psi.east)+(15ex, 0.8ex)$) and ($(yaw.west)-(-5.7ex,1.0ex)$) .. ($(yaw.west)+(12.6ex,-1.0ex)$);
\draw [, arrowclr, dashed, dash pattern=on 4pt off 1.7pt] ($(ddlmc.east)-(5.2ex, 1.0ex)$) .. controls ($(ddlmc.east)+(4ex, -1.0ex)$) and ($(output.west)-(5.7ex,0.0ex)$) .. ($(output.west)+(0.5ex,0.0ex)$) node [black, xshift=-5.5ex, yshift=6ex, scale=\inscal]{$[\phi^\star, \theta^\star]$};
\draw [->, arrowclr, dashed, dash pattern=on 4pt off 1.7pt] ($(tmaf.east)-(5.5ex, 0ex)$) .. controls ($(tmaf.east)+(4ex, -0.0ex)$) and ($(output.west)-(5.7ex,0.0ex)$) .. ($(output.west)+(0.5ex,0.0ex)$) node [black, xshift=-7ex, yshift=1.2ex, scale=\inscal]{$f^\star$};
\draw [->, arrowclr, dashed, dash pattern=on 4pt off 1.7pt] ($(yaw.east)-(12.5ex, 1ex)$) .. controls ($(yaw.east)+(4ex, 0.0ex)$) and ($(output.west)-(5.7ex,0.0ex)$) .. ($(output.west)+(0.5ex,0.0ex)$) node [black, xshift=-5.5ex, yshift=-6.8ex, scale=\inscal]{$\psi^\star / \omega^\star_z$};
\draw [, white!0!black, dashed, dash pattern=on 4pt off 1.7pt] ($(yaw.west)+(3.5ex, 0.8ex)$) .. controls ($(yaw.west)+(10ex, -4.3ex)$) and ($(yaw.east)-(10.7ex,4.3ex)$) .. ($(yaw.east)-(3.5ex,-0.6ex)$);
}};

\end{tikzpicture}
\vspace{-2.8ex}
%
\caption{High-level design of \OURS.}
\label{fig:control}
\vspace{-3.0ex}
\end{figure}
%

%
\section{\OURSFull}
\label{sec:\OURS}
In UAVs, the position control is developed specifically for a problem under consideration. In the same way, we also develop an end-to-end control design \OURS{} (Fig.~\ref{fig:control}) for off-center aerial grasping that offers different control modes in different directions, e.g. position control in $\bm z_W$ while velocity control in $\bm x_W$ or $\bm y_W$. It is a crucial functionality to execute aerial grasping operations precisely. \OURS{} achieves that via \texttt{TMAF} and Decoupled Motion Control (\texttt{DMC}) to generate $\{f_B^\star, \phi^\star, \theta^\star, \psi^\star\}$ precisely from the input needs (different control modes, input state)  instead of Eq.~\ref{eq:classicalthrust} and  \ref{eq:forcedir}.
\par
In this work, we have focused on PID-based loops, which can be replaced by any other controller as future improvements, e.g. non-linear control, such as a robust or sliding mode, etc. 
\subsection{Position Feedback Loop}
\label{sec:positionloop}
The position feedback loop of \OURS{} produces the $\bm a^\star_B$ commands. It has two cascaded controllers, motivated by the need for position control in aerial grasping to perform precise navigation (via position commands) and visual servoing (via velocity commands) during grasping operations.
\subsubsection{Position Controller}
It outputs virtual velocity commands $\bm v^\star_B$, and the control law is defined as follows:
 \begin{equation}
 \footnotesize
 \textbf{PID}_{\bm p} \equiv \bm v^\star_B = \bm K^p_p \bm e_{\bm p} + \bm K^p_i \int {\bm e}_{\bm p} dt +  \bm K^p_d \bm \dot{\bm e}_{\bm p}
  \label{eq:posctrl}
 \end{equation}
where, $\bm e_{\bm p} = \bm p^\star_B - \bm p_B$ is position error, $\bm K^p \equiv \{\bm K^p_p, \bm K^p_i, \bm K^p_d\}$. $\bm K^p_p = \text{diag}(k^p_{p_x}, k^p_{p_y}, k^p_{p_z})$, $\bm K^p_i = \text{diag}(k^p_{i_x}, k^p_{i_y}, k^p_{i_z})$, and $\bm K^p_d = \text{diag}(k^p_{d_x}, k^p_{d_y}, k^p_{d_z})$ are PID gains in the $\{\bm x, \bm y, \bm z\}$ axes.
\subsubsection{Velocity Controller}
It generates virtual commands based on $\bm v^\star_B$ such that $\bm v_B \rightarrow \bm v^\star_B$. The control law is as follows:
 \begin{equation}
 \footnotesize
\textbf{PID}_{\bm v} \equiv  \bm q_B = \bm K^v_p \bm e_{\bm v} + \bm K^v_i \int {\bm e}_{\bm v} dt + \bm K^v_d \bm \dot{\bm e}_{\bm v}
 \label{eq:velctrl}
 \end{equation}
where $\bm e_{\bm v} = \bm v^\star_B - \bm v_B$ is velocity error, $\bm K^v \equiv \{\bm  K^v_p, \bm  K^v_i, \bm K^v_d\}$. $\bm K^v_p = \text{diag}(k^v_{p_x}, k^v_{p_y}, k^v_{p_z})$, $\bm K^v_i = \text{diag}(k^v_{i_x}, k^v_{i_y}, k^v_{i_z})$, and $\bm K^v_d = \text{diag}(k^v_{d_x}, k^v_{d_y}, k^v_{d_z})$ are the PID gains. 
%
%
%
\par
The output of $\textbf{PID}_{\bm v}$ i.e. $\bm q_B=\{q_x, q_y, q_z\} \equiv \bm a^\star_B$ is fed to \texttt{TMAF} which produces the desired thrust intensity and direction i.e. $\{f_B^\star, \phi^\star, \theta^\star, \psi^\star\}$ similar to MI or DA.
\par
In this paper, we are also interested in decoupled control of quadrotor, which utilizes \texttt{TMAF} only in the $z$ direction and utilizes the output of $\textbf{PID}_{\bm v}$ differently, as discussed below. 

\subsection{Decoupled Motion Controller (\texttt{DMC})}
\label{sec:ddlmc}
Inspired by the $2$D motion, we propose a decoupled control of the quadrotor which boils down to controlling the vertical direction ($\{z, v_z, a_z\}$), horizontal direction ($\{x,y,v_x, v_y, a_x, a_y\}$), and heading ($\psi$) independently. Since $\{\phi, \theta\}$ governs the horizontal motion i.e. in $\{x,y\}$, we assume the quadrotor as a $2$D-agent navigating in $xy$-plane with $z$ controlled by \texttt{TMAF}, and $x,y$ by \texttt{DMC}, with heading control done separately.
\par
In \texttt{DMC}, we propose to compute $\{\phi^\star, \theta^\star\}$ directly without using forces. This is in contrast to the existing works, such as geometric tracking controller \cite{geometrictracking} that uses forces in all directions to obtain $\{\phi^\star, \theta^\star\}$, whereas \texttt{DMC} uses force only in $z$ direction. This design helps \texttt{DMC} to counter unmodeled disturbances induced in the lateral directions caused by the off-centered dynamic payloads. This approach is advantageous for aerial grasping to develop complex state machines \cite{towards}. 
\par
\subsubsection{Direction $\{\phi^\star, \theta^\star\}$} \texttt{DMC} has two independent controllers, one for each of $\bm x_W$ and $\bm y_W$, producing $\{\phi^\star, \theta^\star\}$. Since $\phi, \theta$ induce motion in $x_B$ and $y_B$, the $\{v^\star_x, v^\star_y\}$ need to be mapped into the body frame due to the presence of heading ($\psi$). We denote the body frame velocity as $v_D$ which is fed to $\textbf{PID}_{\bm v}$ to obtain $\bm q=\{q_{x}, q_{y}\}$ that essentially represents $\{\phi^\star, \theta^\star\}$. The above operation is summarized below:
\begin{equation}
\scriptsize
v^\star_D = \bm R^T_B(\psi) v^\star_B
\end{equation}
\begin{equation}
\footnotesize
[\phi^\star, \theta^\star]^T \equiv \bm q = \mathrm{\mathbf{PID}}(\bm v^\star_D)
\end{equation}
where $\bm R^T_B \in SO2$, transforms $v^\star_B$ from $\mathcal{F}_W$ to  $\mathcal{F}_B$, and $[\phi^\star, \theta^\star]^T$ steers the quadrotor from $[p_x, p_y]$ to $[p^\star_x, p^\star_y]$.
\subsubsection{Thrust Intensity} In \texttt{DMC}, the thrust estimation is done only in $\bm z_W$ to control $\{z, v_z, a_z\}$. It is done via \texttt{TMAF} which outputs $f^\star_{z}$. Then we compute $f^\star_B$ from $f^\star_{z}$ as follows:
\begin{equation}
\footnotesize
f^\star_B = f^\star_{z} / (\bm z_B.\hat{k})
\label{eq:thrust}
\end{equation}
%
%
%
where, $\hat{k}$ is the unit vector along $z_W$. This equation is a simple manipulation of $f^\star_B = \bm z_B^T \bm f^\star_{B}$, given that thrust can only be applied along $\bm z_B$. Here $\bm z_B . \bm z_W =0$ should be avoided, which is generally done by limiting the angles in this range, otherwise $f_z$ becomes zero since $\phi$ or $\theta = 90^\text{o}$, thus causing free-fall of the quadrotor, unless $\phi$ and $\theta$ ensure $\bm z_B . \bm z_W \neq 0$.
\subsubsection{Yaw Controller}
\label{sec:yawcontrol}
GT \cite{geometrictracking} uses the heading $\psi^\star$ (obtained from the user or planner) to calculate  $\{ \phi^\star, \theta^\star \}$. Whereas, we control $\psi^\star$ independently via a PID controller that produces $\omega_z^\star$ based on $\psi^*$. The control law is given as:
\begin{equation}
\footnotesize
\textbf{PID}_{\omega} \equiv  \omega_z^\star = k_p  e_{\psi} + k_i \int e_\psi dt + k_d \dot{e}_{\psi}
\label{eq:yaw}
\end{equation}
where, $e_{\psi} = \psi^\star - \psi$. The $\omega^\star_z$ is sent to the angular rate controller to achieve $\psi \rightarrow \psi^\star$. If required, $\psi^\star$ can also be sent to the attitude controller. 
%
%
%
%
%
\begin{figure}[t]
\centering
\begin{tikzpicture}

\colorlet{posdclr}{white!0!green}
\colorlet{veldclr}{white!0!cyan}
\colorlet{accjdclr}{white!0!magenta}

\colorlet{slamdclr}{white!0!yellow}
\colorlet{slamclr}{white!90!black}

\colorlet{ekfdclr}{white!0!blue}
\colorlet{ekfclr}{white!90!black}

\colorlet{adpddclr}{green!50!black}
\colorlet{adpdclr}{white!90!black}

\colorlet{attdclr}{white!90!black}
\colorlet{attclr}{white!90!black}

\colorlet{mxrdclr}{white!90!black}
\colorlet{mxrclr}{white!90!black}

\colorlet{uavdclr}{white!90!black}
\colorlet{uavclr}{white!90!black}

\colorlet{drawclr}{white!40!black}

\node [scale = 0.60]
{
\tikz{
\node (outer) [draw=white!50!gray, rounded corners=0.3ex, minimum width=90ex, minimum height=24.0ex, xshift=0ex, yshift=-1.5ex]{};
\node (uav) [draw=uavdclr, rounded corners=0.3ex, xshift=36ex, yshift=-1.0ex]{\includegraphics[width=14ex, height=8ex]{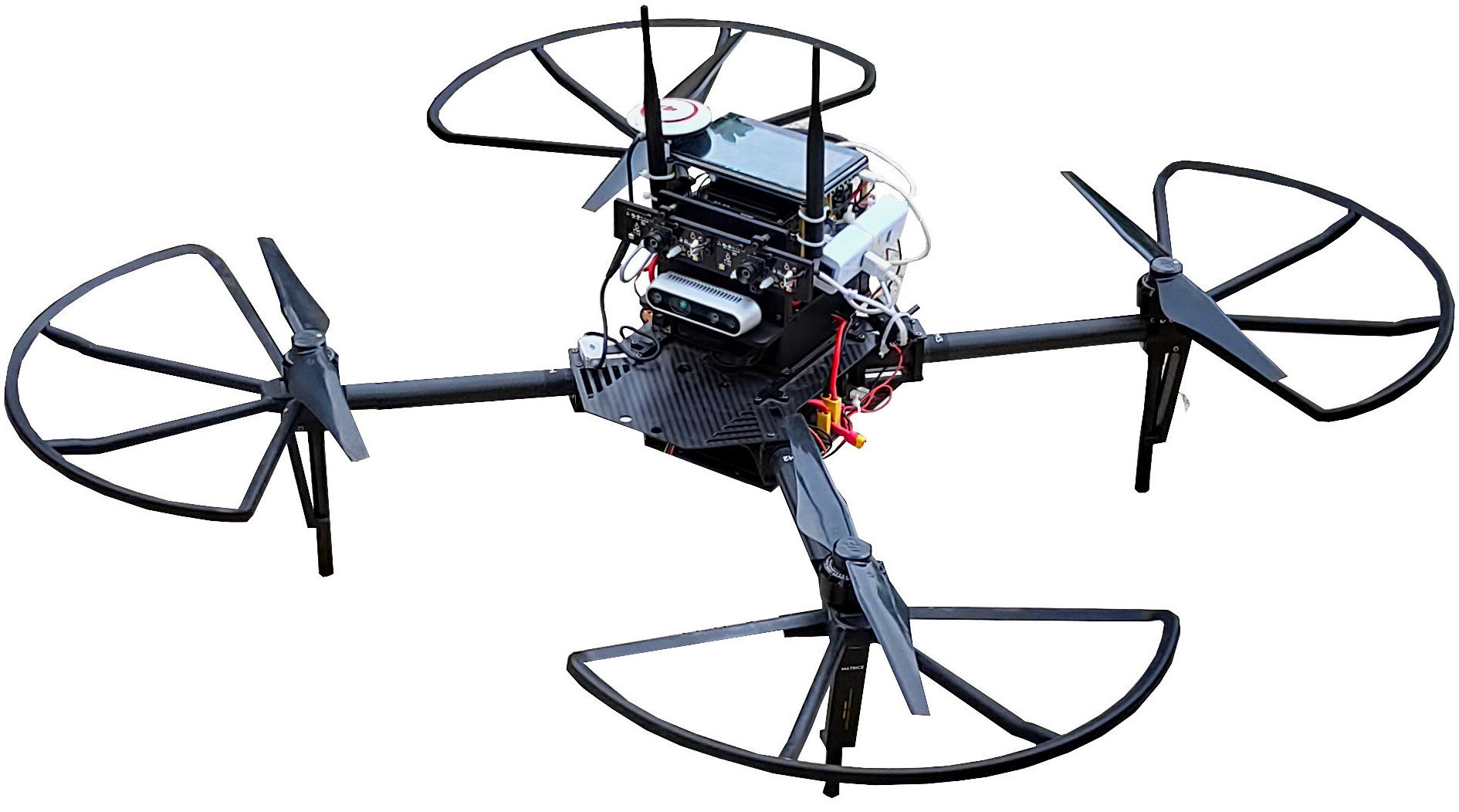}};
\node (vslam) [draw=slamdclr, fill= slamclr, rounded corners=0.3ex, xshift=-33ex, yshift=-10ex]{Visual-SLAM};
\node (ekf) [draw=ekfdclr, fill= ekfclr, rounded corners=0.3ex, xshift=-7ex, yshift=-10ex]{EKF};
\node (adpd) [draw=adpddclr, fill= adpdclr, rounded corners=0.3ex, minimum height=7ex, xshift=-27ex, yshift=5ex]{\shortstack{\OURS \\ Control}};
\node (att) [draw=attdclr, fill= attclr, rounded corners=0.3ex, xshift=6ex, yshift=5ex]{\shortstack{Attitude \\ Control}};
\node (mxr) [draw=mxrdclr, fill= mxrclr, rounded corners=0.3ex, xshift=6ex, yshift=-4.5ex]{\shortstack{Motor \\ Control}};
\draw [->] (vslam) -- (ekf) node[xshift=-11.0ex, yshift=-1.7ex]{$\{\bm p, \bm v, \phi, \theta, \psi\}$};
\draw [->] (ekf) -- ($(ekf.north)+(0ex, 3ex)$) -| (adpd) node[xshift=7.5ex, yshift=-12.0ex]{$\{\bm p, \bm v, \bm a, \bm r\}$};
\draw [->] (adpd) -- (att) node[xshift=-15.5ex, yshift=1.7ex]{$\{\phi^\star, \theta^\star, \psi^\star \text{/} \omega^\star_z, f^\star\}$};
\draw [->] (att) -- (mxr);
\draw [->] (mxr) -- ($(uav.west)-(0ex,3.5ex)$);
\draw [->] (uav) |- (ekf)  node[xshift=14.5ex, yshift=-1.7ex]{$\{\phi, \theta, \psi, \bm a\}$};
\draw [->,draw=posdclr] ($(adpd.west)-(10ex,-2.75ex)$) -- ($(adpd.west)-(0ex,-2.75ex)$) node[xshift=-4.5ex, yshift=1.2ex]{$\bm p^\star$};
\draw [->,draw=veldclr] ($(adpd.west)-(10ex,0.0ex)$) -- ($(adpd.west)-(0ex,0.0ex)$) node[xshift=-4.5ex, yshift=1.2ex]{$\bm v^\star$};
\draw [->,draw=accjdclr] ($(adpd.west)-(10ex,2.75ex)$) -- ($(adpd.west)-(0ex,2.75ex)$) node[xshift=-4.5ex, yshift=1.2ex]{$\psi^\star$};
\draw [->] (uav) |- (att)  node[xshift=10.5ex, yshift=1.7ex]{$\{\phi, \theta, \psi\}$};
}
};
\end{tikzpicture}
\caption{Connectivity between \OURS~and other feedback components.}
\label{fig:fullcontrol}
\vspace{-1ex}
\end{figure}
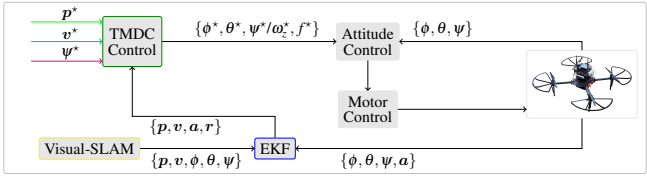

\begin{table}[t]
\centering
\caption{\OURS~gains}
\label{tab:pidgains}
\arrayrulecolor{white!60!black}
\scriptsize
%
%
\setlength{\tabcolsep}{3.1pt}
\vspace{-0.75ex}
\begin{tabular}{c | c | c | c | c}
\hline

 Gain & $\textbf{PID}_{\bm p}$  & $\textbf{PID}_{\bm v}$ & $\texttt{TMAF}$ & $\textbf{PID}_{\omega}$ \\ \hline
 
 $\bm K_p$ & diag($3.0$, $3.0$, $1.0$)   & diag($3.0$, $3.0$, $1.5$) & diag($0.6$, $0.6$, $0.6$) & $0.4$ \\
 $\bm K_i$ & diag($0.1$, $0.1$, $0.1$)   & diag($0.5$, $0.5$, $0.25$) & $-$ & $0.01$ \\
 $\bm K_d$ & diag($0.8$, $0.8$, $0.1$)   & diag($0.5$, $0.5$, $0.20$) & diag($0.02$, $0.02$, $0.02$) & $0.8$ \\
  
   \hline
\end{tabular}
\vspace{-1ex}
\end{table}
\section{Implementation}
\label{sec:impl}
\subsubsection{Loop Rates}
We obtain position feedback via onboard visual-SLAM at $30$Hz, and raw acceleration from the onboard IMU at $80$Hz. They are fused using EKF to estimate the quadrotor state. The $\textbf{PID}_{\bm p}, \textbf{PID}_{\bm v}$ and \texttt{TMAF} runs at $30$Hz, $60$Hz and $80$Hz respectively. The overall information flow between the different controllers is shown in Fig.~\ref{fig:fullcontrol}. 
\subsubsection{Derivative Filtering}
Since derivative controller input can be noisy, we use a cosine-weighted moving average filter over a window of $4$ samples. The weights are given by $w_i=\cos(i\zeta)~\forall~i \in [0,4),~~\zeta=\zeta_{m}/4, \zeta_m=80^o$, and $i=0$ refers to the latest sample. The same applies to the ``jerk'' term in Eq.~\ref{eq:tmafmisctrl} of \texttt{TMAF}, which we compute via Euler backward difference on the accelerometer measurements. 
\subsubsection{Tuning}
 \OURS{} has six PIDs and a \texttt{TMAF} controller, which we tune 
 to obtain a behaviour similar to a critically damped system. We use \texttt{runtime-reconfiguration} and \texttt{rqt\_reconfigure} utility of \texttt{ROS}, which allows changing the gains at runtime, reducing the tuning effort only to $\sim 5-10$ minutes. For tuning, we first act upon PID in $\bm z_W$, then in $\bm x_W$ and $\bm y_W$, which are identical. Due to the cascaded nature of TMDC, we first tune \texttt{TMAF} gains, then $\textbf{PID}_{\bm v}$ and finally $\textbf{PID}_{\bm p}$ and $\textbf{PID}_{\omega}$. Table~\ref{tab:pidgains} shows the tuned values, which can be an initial guess for a different platform. Our \texttt{TMDC} is quite resistant to the variance in its gains. Hence, various controllers of TMDC need not be perfectly tuned to achieve a desirable control performance (Sec.~\ref{sec:varying_gain_effect}).

\input{exp_plots/THL_combined/plot_mini}

  %
 \section{Experiments}
  \label{sec:exp}

\begin{table}[!t]
\centering
\caption{UAV platform specifications.}
\label{tab:specs}
\arrayrulecolor{white!70!black}
\scriptsize
\setlength{\tabcolsep}{1.0pt}
\vspace{-0.75ex}

\begin{tabular}{c c c c}
\hline
\multicolumn{1}{c}{Attribute} & \multicolumn{1}{|c}{Specification} & \multicolumn{1}{||c}{Attribute} & 
\multicolumn{1}{|c}{Specification} \\ \cline{1-4}
\multicolumn{1}{c}{\cfontpt{Size}} & \multicolumn{1}{|c}{\cfontpt{$0.90$m$\times$ $0.90$m$\times$ $0.45$m}} & \multicolumn{1}{||c}{\cfontpt{Operating Voltage}} &  \multicolumn{1}{|c}{\cfontpt{$22$V}} \\
\multicolumn{1}{c}{\cfontpt{Size w/ Gripper}} & \multicolumn{1}{|c}{\cfontpt{$1.5$m $\times$ $0.90$m $\times$ $0.45$m}} & \multicolumn{1}{||c}{\cfontpt{Hover time}} & \multicolumn{1}{|c}{\cfontpt{$20$ minutes}} \\
\multicolumn{1}{c}{\cfontpt{Weight}} & \multicolumn{1}{|c}{\cfontpt{$2.5$Kg}} & \multicolumn{1}{||c}{\cfontpt{Hover time w/ Gripper}} & \multicolumn{1}{|c}{\cfontpt{$15$ minutes}} \\
\multicolumn{1}{c}{\cfontpt{Weight w/ Gripper}} & \multicolumn{1}{|c}{\cfontpt{$3.4$Kg}} & \multicolumn{1}{||c}{\cfontpt{Onboard Computer}} & \multicolumn{1}{|c}{NVIDIA Jetson-NX} \\
\multicolumn{1}{c}{\cfontpt{Rotor diameter}} & \multicolumn{1}{|c}{\cfontpt{$0.46$m}} & \multicolumn{1}{||c}{\cfontpt{Communication}} & \multicolumn{1}{|c}{\cfontpt{UART $@921600$ Baud}} \\
\multicolumn{1}{c}{\cfontpt{Rotor-to-Rotor distance}} & \multicolumn{1}{|c}{\cfontpt{$0.35$m}} & \multicolumn{1}{||c}{\cfontpt{Stereo-Rig}} & \multicolumn{1}{|c}{\cfontpt{$2 \times$~$@432 \times 240, 30$Hz}} \\
 \hline
\end{tabular}
\vspace{-3ex}

\end{table}

We rigorously analyze \OURS~on a quadrotor (Fig.~\ref{fig:m100}, Table~\ref{tab:specs}) in several adverse scenarios: (\textit{i}) different flying workspaces, (\textit{ii}) autonomous take-off, hover, and land, (\textit{iii}) step-response, (\textit{iv}) navigation via trajectory execution, (\textit{v}) battery discharge, (\textit{vi}) non-uniform loop rates and gain variance, and (\textit{vii}) center and off-center dynamic payloads. Finally, we compare \texttt{TMAF} with DA \cite{direct} and \texttt{DMC} with geometric tracking \cite{geometrictracking} comprehensively.  See the video for a real-world demonstration of these experiments.
%

%
\subsection{Constrained Flying Workspaces with Adverse Settings}
We define five workspaces: \texttt{Constrained\_ws}, \texttt{Open\_ws}, \texttt{Open\_ws+Fan}, \texttt{Open\_ws+Gripper}, and \texttt{Open\_ws+Gripper+Payload}. The \texttt{Constrained\_ws} has only $10$cm gap on both sides of the UAV in $\bm y_W$. The \texttt{Open\_ws} has $2.50$m ceiling height and $2$m gap on all sides. 
\par
We analyze the take-off, hover, and land profiles for each workspace to understand \OURS{}'s behaviour (See Fig.~\ref{fig:constrndwsTHL}). Interestingly, regardless of the workspace, \OURS{} exhibits consistent performance for each control input, i.e. $\{x^\star, y^\star, z^\star\}$. The \texttt{Constrained\_ws} poses a challenge in front of \OURS~to stabilize the quadrotor due to a small flying area, intense rotor draft, and turbulence originating due to the large size of the quadrotor. This demands precise thrust controlling. For this reason, for \texttt{Constrained\_ws}, the position feedback loop of TMDC produces noisy desired acceleration ($a^\star_z$), and hence the desired thrust, which indicates frequent corrections. This enables \OURS{} to achieve stable hovering in this workspace while maintaining high accuracy in the lateral direction despite the narrow clearance ($10$cm).
\par
\begin{figure}[t]
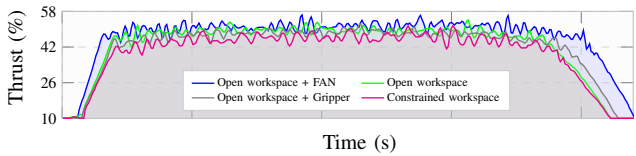

\centering

\begin{tikzpicture}

\FPeval{\xshfta}{0}
\FPeval{\xshftb}{0}
\FPeval{\xshftc}{0}
\FPeval{\xshftd}{0}

\FPeval{\yshfta}{0-0}
\FPeval{\yshftb}{0-9}
\FPeval{\yshftc}{0-18}
\FPeval{\yshftd}{0-27.9}

\FPeval{\pltw}{58}
\FPeval{\plth}{19}

\FPeval{\scal}{1.0}

\FPeval{\mrksize}{0.3}

\colorlet{clr1}{white!0!magenta}
\colorlet{clr2}{white!0!green}
\colorlet{clr3}{white!0!blue}
\colorlet{clr4}{white!0!gray}

\colorlet{gridclr}{white!85!black}
\colorlet{dlegendclr}{white!80!black}
\colorlet{axisclr}{white!75!black}
\colorlet{axisbgclr}{white!99!black}

\colorlet{dlegendclr}{white!80!black}
\colorlet{legendclr}{white!100!black}

\FPeval{\lscale}{0.45}
\FPeval{\limscale}{0.5}
\FPeval{\lxshft}{23.9}
\FPeval{\lyshft}{2.5}

\FPeval{\titlescale}{0.7}
\FPeval{\titlexshift}{21.0}
\FPeval{\titleyshift}{13.7}

\colorlet{gpudclr}{white!80!black}
\colorlet{gpuclr}{white!90!black}
\colorlet{gputxtclr}{white!0!black}

\FPeval{\labelscale}{0.8}
\FPeval{\ticklabelscale}{0.6}

\FPeval{\xlabelxshift}{0+25}
\FPeval{\xlabelyshift}{0+4.5}
\FPeval{\ylabelxshift}{0+4.0}
\FPeval{\ylabelyshift}{0+7.0}

\FPeval{\linew}{0.5}
\FPeval{\dashon}{5.0}
\FPeval{\dashoff}{3.0}

\FPeval{\xshfttakeoff}{0-16}
\FPeval{\xshfthover}{0}
\FPeval{\xshftland}{16}
\FPeval{\opacty}{0.5}
\FPeval{\whitemx}{80}
\node (a) [ xshift = \xshfta ex, yshift=\yshfta ex, scale=\scal]{
\tikz{
\pgfplotsset{width=\pltw ex, height=\plth ex}
\begin{axis}[
   axis background style={fill=axisbgclr},
    title={},
    xlabel={Time (s)},
    ylabel={ Thrust ($\%$)},
    xmin=0, xmax=0.88,
    ymin=10, ymax=58,
    xtick={0, 0.20, 0.4, 0.6, 0.8, 1},
    ytick={10, 26, 42, 58},
    xticklabels={},
    yticklabels={$10$, $26$, $42$, $58$},
     axis line style={axisclr},
    legend image post style={scale =\limscale},
    legend style={at={(\lxshft ex,\lyshft ex)},anchor=south, legend columns = 2, draw = {dlegendclr}, fill={legendclr}, nodes={scale=\lscale}},
    ymajorgrids=true, 
    xmajorgrids=true,
    grid style={dashed, gridclr},
    major tick length=1ex,
    x label style={at={(\xlabelxshift ex, \xlabelyshift ex)},scale=\labelscale},
    y label style={at={(\ylabelxshift ex, \ylabelyshift ex)},scale=\labelscale},
    xticklabel style={scale=\ticklabelscale},
    yticklabel style={scale=\ticklabelscale},
    legend cell align={left}
]
\FPeval{\opacty}{0.2}
\input{exp_plots/thrust_all_ws/th_open_ws_fan}
\input{exp_plots/thrust_all_ws/th_open_ws_fan_no_fill}
\input{exp_plots/thrust_all_ws/th_open_ws}
\input{exp_plots/thrust_all_ws/th_open_ws_no_fill}
\input{exp_plots/thrust_all_ws/th_open_ws_gripper}
\input{exp_plots/thrust_all_ws/th_open_ws_gripper_no_fill}
\input{exp_plots/thrust_all_ws/th_constrained_ws}
\input{exp_plots/thrust_all_ws/th_constrained_ws_no_fill}
%
\legend{Open workspace + FAN\text{~~}, Open workspace\text{~~}, Open workspace + Gripper\text{~~}, Constrained workspace};
  %
\end{axis}
%
}
};

\end{tikzpicture}
\vspace{-1.5ex}
\caption{Thrust profile of TMDC in various workspaces.}
\label{fig:thrustprofileallws}
\vspace{-1ex}
\end{figure}

Similarly,  in \texttt{Open\_ws+Fan}, the \texttt{TMAF} adjusts the thrust to counter the external disturbances precisely while maintaining high hovering accuracy. This workspace has the highest thrust needed because the overhead fan tries to push the UAV downwards (Fig.~\ref{fig:thrustprofileallws}). The other two workspaces have relatively lower aggressive microstepping due to eased constraints.
\begin{table}[!t]
\centering
\caption{Takeoff-Hover-Land accuracy, average of $10$ trails.}
\label{tab:thlperformance}
\arrayrulecolor{white!70!black}
\scriptsize
\setlength{\tabcolsep}{5.0pt}
\vspace{-0.25ex}

\begin{tabular}{c c c c c c c c}
\hline
\multicolumn{1}{c}{\multirow{3}{*}{Workspace}} & \multicolumn{7}{|c}{RMSE (m)} \\ \cline{2-8}
& \multicolumn{2}{|c}{Take-off} & \multicolumn{3}{|c}{Hover} & \multicolumn{2}{|c}{Land} \\  \cline{2-8}

& \multicolumn{1}{|c}{$x$} & $y$ & \multicolumn{1}{|c}{$x$} & $y$ & $z$ & \multicolumn{1}{|c}{$x$} & $y$ \\ \hline
\cfontpt{\texttt{Constrained\_ws}}               & \multicolumn{1}{|c}{$0.04$} & $0.03$ & \multicolumn{1}{|c}{$0.04$} & $0.03$ & $0.02$  & \multicolumn{1}{|c}{$0.03$}  & $0.04$ \\
\cfontpt{\texttt{Open\_ws}}                      & \multicolumn{1}{|c}{$0.06$} & $0.06$ & \multicolumn{1}{|c}{$0.03$} & $0.05$ & $0.02$  & \multicolumn{1}{|c}{$0.03$}  & $0.05$ \\
\cfontpt{\texttt{Open\_ws+Fan}}                & \multicolumn{1}{|c}{$0.03$} & $0.02$ & \multicolumn{1}{|c}{$0.02$} & $0.07$ & $0.02$  & \multicolumn{1}{|c}{$0.03$}  & $0.05$  \\
\cfontpt{\texttt{Open\_ws+Gripper}}            & \multicolumn{1}{|c}{$0.06$} & $0.07$ & \multicolumn{1}{|c}{$0.02$} & $0.06$ & $0.02$  & \multicolumn{1}{|c}{$0.03$}  & $0.09$  \\
\cfontpt{\texttt{Open\_ws+Gripper+Payload}} & \multicolumn{1}{|c}{$0.07$} & $0.08$ & \multicolumn{1}{|c}{$0.03$} & $0.07$ & $0.02$  & \multicolumn{1}{|c}{$0.02$}  & $0.07$ \\
 \hline
\end{tabular}
\vspace{-3.5ex}

\end{table}
\subsection{Takeoff--Hover--Landing Quantitative Performance}
Based on \OURS, we also develop high-level controllers for smooth takeoff and landing, crucial for flight autonomy, while hovering is achieved directly via requesting the set-point.
\subsubsection{Takeoff}
For takeoff, we use position control in $\bm x_W, \bm y_W$ and velocity control in $\bm z_W$ with $x^\star=y^\star=0$ and $v^\star_z=0.20$m/s, allowing precise vertical takeoff without being affected by the ground-effects \cite{groundceiling} (see Sec.~\ref{sec:ground_effect}). From Fig.~\ref{fig:constrndwsTHL}, it is noticeable that \OURS~exhibits the desired response, owing to \texttt{TMAF} in $z$ and \texttt{DMC} in $x-y$, and visible in Table~\ref{tab:thlperformance}.
\subsubsection{Hovering}
We use position control in all the axes with $x^\star=y^\star=0$, $z^\star=0.5$m. Fig.~\ref{fig:constrndwsTHL} indicates a highly accurate hovering response of \OURS, also verifiable via Table~\ref{tab:thlperformance}.
\subsubsection{Landing}
Similar to takeoff, we set $x^\star=y^\star=0$, $v^\star_z=-0.10$m/s. Fig.~\ref{fig:constrndwsTHL} indicates a small error between landing and takeoff position, which is a desirable attribute to execute landing in constrained workspaces precisely (Table~\ref{tab:thlperformance}).

\subsection{Step Response}
%
%


\begin{figure}[t]
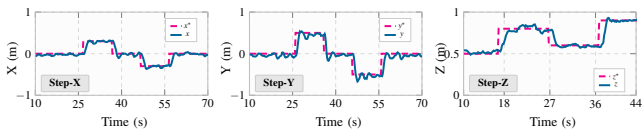

\centering

\begin{tikzpicture}

\FPeval{\xshfta}{0}
\FPeval{\xshftb}{18}
\FPeval{\xshftc}{36}

\FPeval{\yshfta}{0-0}
\FPeval{\yshftb}{0-0}
\FPeval{\yshftc}{0-0}

\FPeval{\pltw}{39}
\FPeval{\plth}{24}

\FPeval{\scal}{0.5}

\FPeval{\mrksize}{0.3}

\colorlet{desclr}{white!0!magenta}
\colorlet{currclr}{blue!60!green}

\colorlet{gridclr}{white!85!black}
\colorlet{dlegendclr}{white!80!black}
\colorlet{axisclr}{white!75!black}
\colorlet{axisbgclr}{white!99!black}

\colorlet{dlegendclr}{white!80!black}
\colorlet{legendclr}{white!100!black}

\FPeval{\lscale}{0.6}
\FPeval{\limscale}{0.5}
\FPeval{\lxshft}{24.5}
\FPeval{\lyshft}{9.0}
\FPeval{\lxshftc}{24.5}
\FPeval{\lyshftc}{0.55}

\FPeval{\titlescale}{0.8}
\FPeval{\titlexshift}{5.0}
\FPeval{\titleyshift}{2.0}

\colorlet{gpudclr}{white!80!black}
\colorlet{gpuclr}{white!90!black}
\colorlet{gputxtclr}{white!0!black}

\FPeval{\labelscale}{1.0}
\FPeval{\ticklabelscale}{0.9}

\FPeval{\xlabelxshift}{0+15}
\FPeval{\xlabelyshift}{0+0.5}
\FPeval{\ylabelxshift}{0+4.0}
\FPeval{\ylabelyshift}{0+6.2}

\FPeval{\linew}{1.5}
\FPeval{\dashon}{5.0}
\FPeval{\dashoff}{3.0}

\FPeval{\xshfttakeoff}{0-16}
\FPeval{\xshfthover}{0}
\FPeval{\xshftland}{16}
\node (a) [xshift = \xshfta ex, yshift=\yshfta ex, scale=\scal]{
\tikz{
\pgfplotsset{width=\pltw ex, height=\plth ex}
\begin{axis}[
   axis background style={fill=axisbgclr},
    title={},
    xlabel={Time (s)},
    ylabel={ X (m)},
    xmin=10, xmax=70,
    ymin=-1, ymax=1,
    xtick={10, 25, 40, 55, 70},
    ytick={-1, 0, 1},
    xticklabels={$10$, $25$, $40$, $55$, $70$},
    yticklabels={$-1$, $0$, $1$},
     axis line style={axisclr},
    legend image post style={scale =\limscale},
    legend style={at={(\lxshft ex,\lyshft ex)},anchor=south, legend columns = 1, draw = {dlegendclr}, fill={legendclr}, nodes={scale=\lscale}},
    ymajorgrids=true, 
    xmajorgrids=true,
    grid style={dashed, gridclr},
    major tick length=1ex,
    x label style={at={(\xlabelxshift ex, \xlabelyshift ex)},scale=\labelscale},
    y label style={at={(\ylabelxshift ex, \ylabelyshift ex)},scale=\labelscale},
    xticklabel style={scale=\ticklabelscale},
    yticklabel style={scale=\ticklabelscale},
]
\input{exp_plots/step_responses/x_des}
\input{exp_plots/step_responses/x}
  \legend{$x^\star$, $x$}
\end{axis}
\node [draw=gpudclr,fill=gpuclr,rounded corners=0.2ex, minimum width=10.0ex, xshift=\titlexshift ex, yshift=\titleyshift ex, scale=\titlescale]{\textcolor{gputxtclr}{\textbf{Step-X}}};
}};
\node (b) [xshift = \xshftb ex, yshift=\yshftb ex, scale=\scal]{
\tikz{
\pgfplotsset{width=\pltw ex, height=\plth ex}
\begin{axis}[
   axis background style={fill=axisbgclr},
    title={},
    xlabel={Time (s)},
    ylabel={ Y (m)},
    xmin=10, xmax=70,
    ymin=-1, ymax=1,
    xtick={10, 25, 40, 55, 70},
    ytick={-1, 0, 1},
    xticklabels={$10$, $25$, $40$, $55$, $70$},
    yticklabels={$-1$, $0$, $1$},
     axis line style={axisclr},
    legend image post style={scale =\limscale},
    legend style={at={(\lxshft ex,\lyshft ex)},anchor=south, legend columns = 1, draw = {dlegendclr}, fill={legendclr}, nodes={scale=\lscale}},
    ymajorgrids=true, 
    xmajorgrids=true,
    grid style={dashed, gridclr},
    major tick length=1ex,
    x label style={at={(\xlabelxshift ex, \xlabelyshift ex)},scale=\labelscale},
    y label style={at={(\ylabelxshift ex, \ylabelyshift ex)},scale=\labelscale},
    xticklabel style={scale=\ticklabelscale},
    yticklabel style={scale=\ticklabelscale},
]
\input{exp_plots/step_responses/y_des}
\input{exp_plots/step_responses/y}
  \legend{$y^\star$, $y$} %
\end{axis}
\node [draw=gpudclr,fill=gpuclr,rounded corners=0.2ex, minimum width=10.0ex, xshift=\titlexshift ex, yshift=\titleyshift ex, scale=\titlescale]{\textcolor{gputxtclr}{\textbf{Step-Y}}};
}};
\node (c) [xshift = \xshftc ex, yshift=\yshftc ex, scale=\scal]{
\tikz{
\pgfplotsset{width=\pltw ex, height=\plth ex}
\begin{axis}[
   axis background style={fill=axisbgclr},
    title={},
    xlabel={Time (s)},
    ylabel={ Z (m)},
    xmin=10, xmax=44,
    ymin=-0.0, ymax=1.0,
    xtick={10, 18, 27, 36, 44},
    ytick={0, 0.5, 1},
    xticklabels={$10$, $18$, $27$, $36$, $44$},
    yticklabels={$0$, $0.5$, $1$},
     axis line style={axisclr},
    legend image post style={scale =\limscale},
    legend style={at={(\lxshftc ex,\lyshftc ex)},anchor=south, legend columns = 1, draw = {dlegendclr}, fill={legendclr}, nodes={scale=\lscale}},
    ymajorgrids=true, 
    xmajorgrids=true,
    grid style={dashed, gridclr},
    major tick length=1ex,
    x label style={at={(\xlabelxshift ex, \xlabelyshift ex)},scale=\labelscale},
    y label style={at={(\ylabelxshift ex, \ylabelyshift ex)},scale=\labelscale},
    xticklabel style={scale=\ticklabelscale},
    yticklabel style={scale=\ticklabelscale},
]
\input{exp_plots/step_responses/z_des}
\input{exp_plots/step_responses/z}
  \legend{$z^\star$, $z$}
\end{axis}
\node [draw=gpudclr,fill=gpuclr,rounded corners=0.2ex, minimum width=10.0ex, xshift=\titlexshift ex, yshift=\titleyshift ex, scale=\titlescale]{\textcolor{gputxtclr}{\textbf{Step-Z}}};
}};
\end{tikzpicture}
\vspace{-1.5ex}
\caption{Step responses in $\mathcal{F}_W$, conducted as different experiments.}
\label{fig:stepresponse}
\vspace{-2ex}
\end{figure}
We also analyse the step response of \OURS~in all directions (Fig.~\ref{fig:stepresponse}). It can be seen that \OURS~converges quickly to the new set-points, and stays there precisely. A slight overshoot is also visible during set-point changes, which was caused to counter heavy turbulence indoors due to the large UAV size.
\subsection{Trajectory Execution or Non-Stationary Configuration}
\begin{figure}[!t]
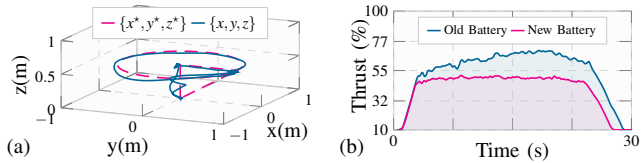

\centering

\begin{tikzpicture}

\FPeval{\xshfta}{0}
\FPeval{\xshftb}{28}

\FPeval{\yshfta}{0-0}
\FPeval{\yshftb}{0-0}

\FPeval{\pltw}{30}
\FPeval{\plth}{20}

\FPeval{\scal}{1.0}

\FPeval{\mrksize}{0.3}

\colorlet{desclr}{white!0!magenta}
\colorlet{currclr}{blue!60!green}

\colorlet{gridclr}{white!85!black}
\colorlet{dlegendclr}{white!80!black}
\colorlet{axisclr}{white!75!black}
\colorlet{axisbgclr}{white!99!black}

\colorlet{dlegendclr}{white!80!black}
\colorlet{legendclr}{white!100!black}

\FPeval{\lscale}{0.6}
\FPeval{\lscaleb}{0.5}
\FPeval{\limscale}{0.3}

\FPeval{\lxshft}{10}
\FPeval{\lyshft}{0+5.8}

\FPeval{\lxshftb}{10.0}
\FPeval{\lyshftb}{8.3}

\FPeval{\titlescale}{0.7}
\FPeval{\titlexshift}{21.0}
\FPeval{\titleyshift}{13.7}

\colorlet{gpudclr}{white!80!black}
\colorlet{gpuclr}{white!90!black}
\colorlet{gputxtclr}{white!0!black}

\FPeval{\labelscale}{0.8}
\FPeval{\ticklabelscale}{0.6}

\FPeval{\xlabelxshiftb}{0+10}
\FPeval{\xlabelyshiftb}{0+4.0}
\FPeval{\ylabelxshiftb}{0+5.0}
\FPeval{\ylabelyshiftb}{0+6.5}

\FPeval{\xlabelxshift}{0+20}
\FPeval{\xlabelyshift}{0-0.5}
\FPeval{\ylabelxshift}{0+4}
\FPeval{\ylabelyshift}{0-1.5}
\FPeval{\zlabelxshift}{0-2.0}
\FPeval{\zlabelyshift}{0+3.0}

\FPeval{\linew}{0.6}
\FPeval{\dashon}{5.0}
\FPeval{\dashoff}{3.0}

\FPeval{\xshfttakeoff}{0-16}
\FPeval{\xshfthover}{0}
\FPeval{\xshftland}{16}
\node (na) [draw=none, xshift =\xshfta ex, yshift= \yshfta ex, scale=\scal]{
\tikz{
\pgfplotsset{width=\pltw ex, height=\plth ex}
\begin{axis}[
   axis background style={fill=axisbgclr},
    title={},
    xlabel={x(m)},
    ylabel={y(m)},
    zlabel={z(m)},
    xmin=-1.0, xmax=1.0,
    ymin=-1.0, ymax=1.0,
    zmin=0, zmax=1.0,
    xtick={-1, 0, 1},
    ytick={-1, 0, 1},
     axis line style={axisclr},
    legend image post style={scale =\limscale},
    legend style={at={(\lxshft ex,\lyshft ex)},anchor=south, legend columns = 2, draw = {dlegendclr}, fill={legendclr}, nodes={scale=\lscale}},
    ymajorgrids=true, 
    xmajorgrids=true,
    zmajorgrids=true, 
    grid style={dashed, gridclr},
    major tick length=1ex,
    x label style={at={(\xlabelxshift ex, \xlabelyshift ex)},scale=\labelscale},
    y label style={at={(\ylabelxshift ex, \ylabelyshift ex)},scale=\labelscale},
    z label style={at={(\zlabelxshift ex, \zlabelyshift ex)},scale=\labelscale},
    xticklabel style={scale=\ticklabelscale},
    yticklabel style={scale=\ticklabelscale},
    zticklabel style={scale=\ticklabelscale},
    legend cell align={left},
]
\input{exp_plots/circle_trajectory/des_traj}
%
%
\input{exp_plots/circle_trajectory/executed_traj}
 \legend{{$\{x^\star, y^\star, z^\star \}$},  {$\{ x, y, z \}$}}
\end{axis}
}};
\node (b) [xshift = \xshftb ex, yshift=\yshftb ex, scale=\scal]{
\tikz{
\pgfplotsset{width=\pltw ex, height=\plth ex}
\begin{axis}[
   axis background style={fill=axisbgclr},
    title={},
    xlabel={Time (s)},
    ylabel={ Thrust ($\%$)},
    xmin=0, xmax=30,
    ymin=10, ymax=100,
    xtick={0, 30},
    xticklabels={0, 30},
    extra x ticks={0, 7.5, 15, 22.5, 30},
    extra x tick labels={},
    ytick={10, 32, 55, 77, 100},
    yticklabels={$10$, $32$, $55$, $77$, $100$},
     axis line style={axisclr},
    legend image post style={scale =\limscale},
    legend style={at={(\lxshftb ex,\lyshftb ex)},anchor=south, legend columns = 2, draw = {dlegendclr}, fill={legendclr}, nodes={scale=\lscaleb}},
    ymajorgrids=true, 
    xmajorgrids=true,
    grid style={dashed, gridclr},
    major tick length=1ex,
    x label style={at={(\xlabelxshiftb ex, \xlabelyshiftb ex)},scale=\labelscale},
    y label style={at={(\ylabelxshiftb ex, \ylabelyshiftb ex)},scale=\labelscale},
    xticklabel style={scale=\ticklabelscale},
    yticklabel style={scale=\ticklabelscale},
]
\FPeval{\opacty}{0.2}
\input{exp_plots/thrust_new_and_old_battery/th_old}
\input{exp_plots/thrust_new_and_old_battery/th_old_no_fill}
\input{exp_plots/thrust_new_and_old_battery/th_new}
\input{exp_plots/thrust_new_and_old_battery/th_new_no_fill}
  \legend{Old Battery, New Battery}
\end{axis}
%
}};
\node (ta) [below of=a, xshift=-12ex,yshift=1ex]{\footnotesize (a)};
\node (tb) [below of=b, xshift=-12ex,yshift=1ex]{\footnotesize (b)};

\end{tikzpicture}
\subfloat{\label{fig:traj}}
\subfloat{\label{fig:thrustoldnewbattery}}
\vspace{-1.5ex}
\caption{(a) TMDC for trajectory execution $@$radii=$0.7$m, average of $10$ trails, and (b) Thrust output of \texttt{TMAF} for new \textit{vs} old battery.}
\vspace{-2ex}
\end{figure}
We also evaluate \OURS~for trajectory execution. Fig.~\ref{fig:traj} shows circular trajectory execution response, averaged over $10$ trails; $5$ with gripper and $5$ without gripper. This analysis shows that apart from hovering, \texttt{TMAF} outputs accurate thrust even during navigation, trajectory execution or non-stationary configuration while handling unknown payloads. 
%
%
%
%
\subsection{Battery Discharge \& Battery Life}
We analyze \texttt{TMAF} behaviour for two fully-charged batteries: a new, and an old one having decreased current supplying capability and heating issues. Fig.~\ref{fig:thrustoldnewbattery} shows our analysis. 
\par
The high thrust value in the case of the old battery indicates that motors need more power to generate a thrust equivalent to the new battery because the old battery is incapable of supplying as much power as supplied by the newer one for the same amount of thrust. \texttt{TMAF} handles this issue easily via microstepped thrust updates (Eq.~\ref{eq:thrusttmaf}), enabling robustness in \OURS{} against novel unmodeled situations. 
%
%
\begin{figure}[!t]
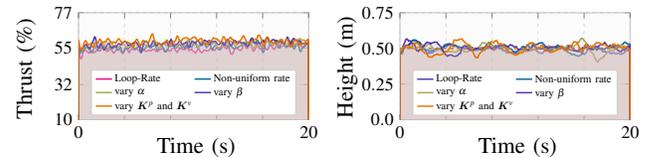

\centering

\begin{tikzpicture}

\FPeval{\xshfta}{0}
\FPeval{\xshftb}{27}
\FPeval{\xshftc}{0}
\FPeval{\xshftd}{0}

\FPeval{\yshfta}{0-0}
\FPeval{\yshftb}{0-0}
\FPeval{\yshftc}{0-18}
\FPeval{\yshftd}{0-27.9}

\FPeval{\pltw}{53}
\FPeval{\plth}{30}

\FPeval{\scal}{0.45}

\FPeval{\mrksize}{0.3}

\colorlet{clr1}{white!0!magenta}
\colorlet{clr2}{blue!60!green}
\colorlet{clr3}{black!40!yellow}
\colorlet{clr4}{blue!60!brown}
\colorlet{clr5}{black!10!orange}

\colorlet{gridclr}{white!85!black}
\colorlet{dlegendclr}{white!80!black}
\colorlet{axisclr}{white!75!black}
\colorlet{axisbgclr}{white!99!black}

\colorlet{dlegendclr}{white!80!black}
\colorlet{legendclr}{white!100!black}
\colorlet{desclr}{white!0!brown}

\FPeval{\lscale}{0.9}
\FPeval{\limscale}{0.8}

\FPeval{\lxshfta}{21.5}
\FPeval{\lxshftb}{21.5}

\FPeval{\lyshfta}{3.8}
\FPeval{\lyshftb}{0.8}

\FPeval{\titlescale}{0.7}
\FPeval{\titlexshift}{21.0}
\FPeval{\titleyshift}{13.7}

\colorlet{gpudclr}{white!80!black}
\colorlet{gpuclr}{white!90!black}
\colorlet{gputxtclr}{white!0!black}

\FPeval{\labelscale}{1.9}
\FPeval{\ticklabelscale}{1.5}

\FPeval{\xlabelxshift}{0+22}
\FPeval{\xlabelyshift}{0+4.5}
\FPeval{\ylabelxshift}{0-2.0}
\FPeval{\ylabelyshift}{0+12.5}

\FPeval{\xlabelxshiftb}{0+22}
\FPeval{\xlabelyshiftb}{0+1.5}
\FPeval{\ylabelxshiftb}{0-2.0}
\FPeval{\ylabelyshiftb}{0+9.9}

\FPeval{\linew}{1.2}
\FPeval{\dashon}{5.0}
\FPeval{\dashoff}{3.0}

\FPeval{\xshfttakeoff}{0-16}
\FPeval{\xshfthover}{0}
\FPeval{\xshftland}{16}

\node (a) [xshift = \xshfta ex, yshift=\yshfta ex, scale=1.0]{
\scalebox{\scal}
{
\tikz{
\pgfplotsset{width=\pltw ex, height=\plth ex}
\begin{axis}[
   axis background style={fill=axisbgclr},
    title={},
    xlabel={Time (s)},
    ylabel={ Thrust ($\%$)},
    xmin=0.0, xmax=1.0,
    ymin=10, ymax=77,
    xtick={0, 1},
    xticklabels={0, 20},
    extra x ticks={0, 0.20, 0.40, 0.60, 0.80, 1.00},
    extra x tick labels={},
    ytick={10, 32, 55, 77, 100},
    yticklabels={$10$, $32$, $55$, $77$, $100$},
     axis line style={axisclr},
    legend image post style={scale =\limscale},
    legend style={at={(\lxshfta ex,\lyshfta ex)},anchor=south, legend columns = 2, draw = {dlegendclr}, fill={legendclr}, nodes={scale=\lscale}},
    legend cell align={left},
    ymajorgrids=true, 
    xmajorgrids=true,
    grid style={dashed, gridclr},
    major tick length=1ex,
    x label style={at={(\xlabelxshift ex, \xlabelyshift ex)},scale=\labelscale},
    y label style={at={(\ylabelxshift ex, \ylabelyshift ex)},scale=\labelscale},
    xticklabel style={scale=\ticklabelscale},
    yticklabel style={scale=\ticklabelscale},
]
\FPeval{\opacty}{0.2}
\input{exp_plots/ablations/th_tmaf_1}
\input{exp_plots/ablations/th_tmaf_no_fill_1}
\input{exp_plots/ablations/th_tmaf_2}
\input{exp_plots/ablations/th_tmaf_no_fill_2}
\input{exp_plots/ablations/th_tmaf_3}
\input{exp_plots/ablations/th_tmaf_no_fill_3}
\input{exp_plots/ablations/th_tmaf_4}
\input{exp_plots/ablations/th_tmaf_no_fill_4}
\input{exp_plots/ablations/th_tmaf_5}
\input{exp_plots/ablations/th_tmaf_no_fill_5}
\legend{Loop-Rate, Non-uniform rate, vary $\alpha$, vary $\beta$, vary $\bm K^p$ and $\bm K^v$ }
\end{axis}
%
}
}};

\node (b) [xshift = \xshftb ex, yshift=\yshftb ex, scale=1.0]{
\scalebox{\scal}
{
\tikz{
\pgfplotsset{width=\pltw ex, height=\plth ex}
\begin{axis}[
   axis background style={fill=axisbgclr},
    title={},
    xlabel={Time (s)},
    ylabel={ Height (m)},
    xmin=0.0, xmax=1.0,
    ymin=0, ymax=0.75,
    xtick={0, 1},
    xticklabels={0, 20},
    extra x ticks={0, 0.20, 0.40, 0.60, 0.80, 1.00},
    extra x tick labels={},
    ytick={0.0, 0.25, 0.50, 0.75, 1.0},
    yticklabels={$0.0$, $0.25$, $0.50$, $0.75$, $1.0$},
     axis line style={axisclr},
    legend image post style={scale =\limscale},
    legend style={at={(\lxshftb ex,\lyshftb ex)},anchor=south, legend columns = 2, draw = {dlegendclr}, fill={legendclr}, nodes={scale=\lscale}},
    legend cell align={left},
    ymajorgrids=true, 
    xmajorgrids=true,
    grid style={dashed, gridclr},
    major tick length=1ex,
    x label style={at={(\xlabelxshiftb ex, \xlabelyshiftb ex)},scale=\labelscale},
    y label style={at={(\ylabelxshiftb ex, \ylabelyshiftb ex)},scale=\labelscale},
    xticklabel style={scale=\ticklabelscale},
    yticklabel style={scale=\ticklabelscale},
]
\FPeval{\opacty}{0.2}
\input{exp_plots/ablations/z_tmaf_1}
\input{exp_plots/ablations/z_tmaf_no_fill_1}
\input{exp_plots/ablations/z_tmaf_2}
\input{exp_plots/ablations/z_tmaf_no_fill_2}
\input{exp_plots/ablations/z_tmaf_3}
\input{exp_plots/ablations/z_tmaf_no_fill_3}
\input{exp_plots/ablations/z_tmaf_4}
\input{exp_plots/ablations/z_tmaf_no_fill_4}
\input{exp_plots/ablations/z_tmaf_5}
\input{exp_plots/ablations/z_tmaf_no_fill_5}
\input{exp_plots/ablations/z_des}
\legend{Loop-Rate, Non-uniform rate, vary $\alpha$, vary $\beta$, vary $\bm K^p$ and $\bm K^v$ }
  %
\end{axis}
%
}
}};
\end{tikzpicture}
\vspace{-3.7ex}
\caption{Ablation study of \texttt{TMDC} in various scenarios: (\textit{i}) Scaling loop-rates ($0.33-1.25$), (\textit{ii}) non-uniform loop-rates (a jitter of $\pm5$ ms), (\textit{iii-iv}) scaling $\alpha,  \beta$ of \texttt{TMAF} ($0.75-1.25$), and (\textit{v}) scaling gains of $\textbf{PID}_{\bm p}$, $\textbf{PID}_{\bm v}$  ($0.33-1.25$) in the position feedback loop.}
\label{fig:ablations}
\vspace{-3ex}
\end{figure}
\subsection{Loop Rate Scaling and Non-uniform Loop Rates}
\label{sec:exp_loop_rates}
Fig.~\ref{fig:ablations} shows the analysis. Our observations align with our claims that \texttt{TMDC} can sustain with low frequency and non-uniform execution rates. We have plotted the average value of the experiment conducted in a range of values (Fig.~\ref{fig:ablations}). 
\subsection{Effect of Gain Tuning Variance}
To analyze the effect of gains, we vary the parameters of \texttt{TMAF} i.e. $\alpha$ and $\beta$, and observe the hovering performance. We see that \texttt{TMAF} is robust to the large variance in its gains, establishing that despite the cascaded control structure, TMDC need not be tuned perfectly. See Fig.~\ref{fig:ablations}.
\label{sec:varying_gain_effect}
\input{exp_plots/hover_dyanmic_payload/plot}
\subsection{Payload Conditions}
\label{sec:thldynpayload}
\label{sec:payload}
\subsubsection{Static Off-Center Payload}
Fig.~\ref{fig:thlowsg} shows this experiment. It must be noted that \OURS{} was tuned without the arm, which weighs $900$gm ($36\%$ of the UAV weight). It verifies \OURS{} sustaining large payload variance.
\subsubsection{Dynamic Payload}
Handling large and unknown dynamic payloads is the major accomplishment of \OURS. The Fig.~\ref{fig:hoverdynamicpayload} shows the results. Noticeably, the UAV hovers with high accuracy even after attaching the payloads dynamically.
\par
With a center payload ($700$ gm) and the gripper, the UAV's weight increases by $64\%$, which is quite a large variance. Nonetheless, \OURS{} handles this scenario by increasing the thrust automatically, thanks to the \texttt{TMAF} thrust controller.
\par
In contrast, the effect of off-center payload is much more intense since CoG gets shifted, causing sudden unwanted motions in $\bm x_W$ at the attach and detach events. However, the \texttt{TMAF} and \texttt{DMC} suppress them precisely. Payload analysis with the state-of-the-art controllers is discussed next.
%
%
%
\subsection{\texttt{TMAF} \textit{vs} DA}
\label{sec:exp_tmaf_vs_da}
Due to the code's unavailability, we implement DA \cite{direct}. Both \texttt{TMAF} and DA run at the same frequency (Sec.~\ref{sec:impl}) and are tuned to exhibit their best performance. We analyze the hovering performance of these two controllers in many practical and challenging scenarios, as discussed below. 
\subsubsection{Center and Off-center Dynamic Payloads}
We compare our \texttt{TMAF+DMC} with DA and Geometric Tracking (GT) \cite{geometrictracking} when the gripper is attached to the UAV. We use two weights \texttt{w$1$} and \texttt{w$2$}, and attach them in five different settings (Fig.~\ref{fig:payload_comparison}).
\par
It can be noticed that DA struggles to maintain height during payload attach and detach events. Moreover, it gets severely disturbed in the case of off-center payload, as seen in the $x$-axis performance. Now we use GT in our TMDC, which also has severe displacements in the $x$-axis. On the other hand, our TMDC with the proposed decoupled motion control \texttt{DMC} accurately outputs roll and pitch angles based on the system behaviour in the lateral directions during disturbance to handle off-center payload, as evident from the minimal displacement in $x$, which justifies the importance of decoupled control.
\subsubsection{Extra Heavy Payload}
\begin{figure}[t]
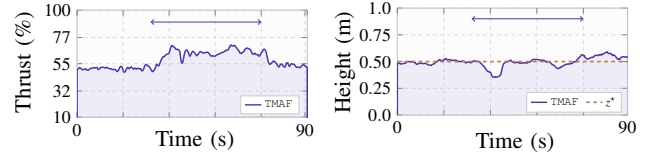

\centering
\colorlet{clr1}{white!0!magenta}
\colorlet{clr2}{blue!60!green}
\colorlet{clr3}{black!40!yellow}
\colorlet{clr4}{blue!60!brown}

\begin{tikzpicture}

\FPeval{\xshfta}{0}
\FPeval{\xshftb}{27}
\FPeval{\xshftc}{0}
\FPeval{\xshftd}{0}

\FPeval{\yshfta}{0-0}
\FPeval{\yshftb}{0-0}
\FPeval{\yshftc}{0-18}
\FPeval{\yshftd}{0-27.9}

\FPeval{\pltw}{53}
\FPeval{\plth}{30}

\FPeval{\scal}{0.45}

\FPeval{\mrksize}{0.3}

\colorlet{gridclr}{white!85!black}
\colorlet{dlegendclr}{white!80!black}
\colorlet{axisclr}{white!75!black}
\colorlet{axisbgclr}{white!99!black}

\colorlet{dlegendclr}{white!80!black}
\colorlet{legendclr}{white!100!black}
\colorlet{desclr}{white!0!brown}

\FPeval{\lscale}{1.0}
\FPeval{\limscale}{0.8}

\FPeval{\lxshfta}{36.0}
\FPeval{\lxshftb}{33.0}

\FPeval{\lyshfta}{2.8}
\FPeval{\lyshftb}{0.5}

\FPeval{\titlescale}{0.7}
\FPeval{\titlexshift}{21.0}
\FPeval{\titleyshift}{13.7}

\colorlet{gpudclr}{white!80!black}
\colorlet{gpuclr}{white!90!black}
\colorlet{gputxtclr}{white!0!black}

\FPeval{\labelscale}{1.9}
\FPeval{\ticklabelscale}{1.5}

\FPeval{\xlabelxshift}{0+22}
\FPeval{\xlabelyshift}{0+4.5}
\FPeval{\ylabelxshift}{0-2.0}
\FPeval{\ylabelyshift}{0+12.5}

\FPeval{\xlabelxshiftb}{0+22}
\FPeval{\xlabelyshiftb}{0+1.5}
\FPeval{\ylabelxshiftb}{0-2.0}
\FPeval{\ylabelyshiftb}{0+9.9}

\FPeval{\linew}{1.2}
\FPeval{\dashon}{5.0}
\FPeval{\dashoff}{3.0}

\FPeval{\xshfttakeoff}{0-16}
\FPeval{\xshfthover}{0}
\FPeval{\xshftland}{16}

\node (a) [xshift = \xshfta ex, yshift=\yshfta ex, scale=1.0]{
\scalebox{\scal}
{
\tikz{
\pgfplotsset{width=\pltw ex, height=\plth ex}
\begin{axis}[
   axis background style={fill=axisbgclr},
    title={},
    xlabel={Time (s)},
    ylabel={ Thrust ($\%$)},
    xmin=0.0, xmax=1.0,
    ymin=10, ymax=100,
    xtick={0,0.99},
    xticklabels={0,90},
    extra x ticks={0, 0.20, 0.40, 0.60, 0.80, 1.00},
    extra x tick labels={},
    ytick={10, 32, 55, 77, 100},
    yticklabels={$10$, $32$, $55$, $77$, $100$},
     axis line style={axisclr},
    legend image post style={scale =\limscale},
    legend style={at={(\lxshfta ex,\lyshfta ex)},anchor=south, legend columns = 4, draw = {dlegendclr}, fill={legendclr}, nodes={scale=\lscale}},
    legend cell align={left},
    ymajorgrids=true, 
    xmajorgrids=true,
    grid style={dashed, gridclr},
    major tick length=1ex,
    x label style={at={(\xlabelxshift ex, \xlabelyshift ex)},scale=\labelscale},
    y label style={at={(\ylabelxshift ex, \ylabelyshift ex)},scale=\labelscale},
    xticklabel style={scale=\ticklabelscale},
    yticklabel style={scale=\ticklabelscale},
    legend cell align={left}
]
\FPeval{\opacty}{0.2}
\input{exp_plots/remote_hanging/th_tmaf}
\input{exp_plots/remote_hanging/th_tmaf_no_fill}
\draw [<->, clr4, thick] (axis cs:0.32,90) -- (axis cs:0.80, 90);
\legend{\texttt{TMAF}}
\end{axis}
%
}
}};

\node (b) [xshift = \xshftb ex, yshift=\yshftb ex, scale=1.0]{
\scalebox{\scal}
{
\tikz{
\pgfplotsset{width=\pltw ex, height=\plth ex}
\begin{axis}[
   axis background style={fill=axisbgclr},
    title={},
    xlabel={Time (s)},
    ylabel={ Height (m)},
    xmin=0.0, xmax=0.99,
    ymin=0, ymax=1.0,
    xtick={0,0.99},
    xticklabels={0,90},
    extra x ticks={0, 0.20, 0.40, 0.60, 0.80, 1.00},
    extra x tick labels={},
    ytick={0.0, 0.25, 0.50, 0.75, 1.0},
    yticklabels={$0.0$, $0.25$, $0.50$, $0.75$, $1.0$},
     axis line style={axisclr},
    legend image post style={scale =\limscale},
    legend style={at={(\lxshftb ex,\lyshftb ex)},anchor=south, legend columns = 4, draw = {dlegendclr}, fill={legendclr}, nodes={scale=\lscale}},
    legend cell align={left},
    ymajorgrids=true, 
    xmajorgrids=true,
    grid style={dashed, gridclr},
    major tick length=1ex,
    x label style={at={(\xlabelxshiftb ex, \xlabelyshiftb ex)},scale=\labelscale},
    y label style={at={(\ylabelxshiftb ex, \ylabelyshiftb ex)},scale=\labelscale},
    xticklabel style={scale=\ticklabelscale},
    yticklabel style={scale=\ticklabelscale},
   legend cell align={left}
]
\FPeval{\opacty}{0.2}
\input{exp_plots/remote_hanging/z_tmaf}
\input{exp_plots/remote_hanging/z_tmaf_no_fill}
\input{exp_plots/remote_hanging/z_des}
\draw [<->, clr4, thick] (axis cs:0.32,0.90) -- (axis cs:0.80,.90);
\legend{\texttt{TMAF}~~,$z^\star$}
\end{axis}
%
}
}};

\end{tikzpicture}
\subfloat{\label{fig:vh_tmaf}} 
\subfloat{\label{fig:vh_tmafz}}
\vspace{-3.7ex}
\newcommand{\harrtmafp}{\raisebox{0.1ex}{\tikz{\draw[<->, clr4] (0,0) -- (0.20,0);}}}
\caption{\OURS{} hovering at very heavy payload of $1.2$Kg and gripper of 0.9Kg. The `\protect\harrtmafp' indicates the payload attached duration.}
\label{fig:remote_hanging}
\vspace{-3ex}
\end{figure}

To go even further, we dynamically attach a heavier weight of $1.2$kg (Fig.~\ref{fig:remote_hanging}), which introduces a variance of $84\%$. Noticeably, \texttt{TMAF} accommodates the new weight seamlessly, as evident from the increased thrust value from $\sim55\%$ to $70\%$. We could easily attach more weight; however, we avoided that to prevent motor heating due to a constant higher power on the motors.
\input{exp_plots/payload_comparison/plot}
\subsubsection{Hovering with Battery Discharge and Off-center Gripper}
\begin{figure}[!t]
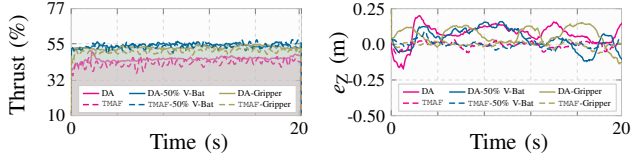

\centering

\begin{tikzpicture}

\FPeval{\xshfta}{0}
\FPeval{\xshftb}{27}
\FPeval{\xshftc}{0}
\FPeval{\xshftd}{0}

\FPeval{\yshfta}{0-0}
\FPeval{\yshftb}{0-0}
\FPeval{\yshftc}{0-18}
\FPeval{\yshftd}{0-27.9}

\FPeval{\pltw}{53}
\FPeval{\plth}{30}

\FPeval{\scal}{0.45}

\FPeval{\mrksize}{0.3}

\colorlet{clr1}{white!0!magenta}
\colorlet{clr2}{blue!60!green}
\colorlet{clr3}{black!40!yellow}
\colorlet{clr4}{blue!60!brown}

\colorlet{gridclr}{white!85!black}
\colorlet{dlegendclr}{white!80!black}
\colorlet{axisclr}{white!75!black}
\colorlet{axisbgclr}{white!99!black}

\colorlet{dlegendclr}{white!80!black}
\colorlet{legendclr}{white!100!black}
\colorlet{desclr}{white!0!brown}

\FPeval{\lscale}{0.76}
\FPeval{\limscale}{0.8}

\FPeval{\lxshfta}{21.7}
\FPeval{\lxshftb}{21.7}

\FPeval{\lyshfta}{3.5}
\FPeval{\lyshftb}{0.8}

\FPeval{\titlescale}{0.7}
\FPeval{\titlexshift}{21.0}
\FPeval{\titleyshift}{13.7}

\colorlet{gpudclr}{white!80!black}
\colorlet{gpuclr}{white!90!black}
\colorlet{gputxtclr}{white!0!black}

\FPeval{\labelscale}{1.9}
\FPeval{\ticklabelscale}{1.5}

\FPeval{\xlabelxshift}{0+22}
\FPeval{\xlabelyshift}{0+4.5}
\FPeval{\ylabelxshift}{0-2.0}
\FPeval{\ylabelyshift}{0+12.5}

\FPeval{\xlabelxshiftb}{0+22}
\FPeval{\xlabelyshiftb}{0+1.5}
\FPeval{\ylabelxshiftb}{0-2.0}
\FPeval{\ylabelyshiftb}{0+9.9}

\FPeval{\linew}{1.2}
\FPeval{\dashon}{5.0}
\FPeval{\dashoff}{3.0}

\FPeval{\xshfttakeoff}{0-16}
\FPeval{\xshfthover}{0}
\FPeval{\xshftland}{16}

\node (a) [xshift = \xshfta ex, yshift=\yshfta ex, scale=1.0]{
\scalebox{\scal}
{
\tikz{
\pgfplotsset{width=\pltw ex, height=\plth ex}
\begin{axis}[
   axis background style={fill=axisbgclr},
    title={},
    xlabel={Time (s)},
    ylabel={ Thrust ($\%$)},
    xmin=0.0, xmax=1.0,
    ymin=10, ymax=77,
    xtick={0,0.99},
    xticklabels={0,20},
    extra x ticks={0, 0.20, 0.40, 0.60, 0.80, 1.00},
    extra x tick labels={},
    ytick={10, 32, 55, 77},
    yticklabels={$10$, $32$, $55$, $77$},
     axis line style={axisclr},
    legend image post style={scale =\limscale},
    legend style={at={(\lxshfta ex,\lyshfta ex)},anchor=south, legend columns = 3, draw = {dlegendclr}, fill={legendclr}, nodes={scale=\lscale}},
    legend cell align={left},
    ymajorgrids=true, 
    xmajorgrids=true,
    grid style={dashed, gridclr},
    major tick length=1ex,
    x label style={at={(\xlabelxshift ex, \xlabelyshift ex)},scale=\labelscale},
    y label style={at={(\ylabelxshift ex, \ylabelyshift ex)},scale=\labelscale},
    xticklabel style={scale=\ticklabelscale},
    yticklabel style={scale=\ticklabelscale},
    legend cell align={left},
]
\FPeval{\opacty}{0.2}
\input{exp_plots/tmaf_vs_da_comparison/th_da}
\input{exp_plots/tmaf_vs_da_comparison/th_da_no_fill}
\input{exp_plots/tmaf_vs_da_comparison/th_da_50}
\input{exp_plots/tmaf_vs_da_comparison/th_da_no_fill_50}
\input{exp_plots/tmaf_vs_da_comparison/th_da_g}
\input{exp_plots/tmaf_vs_da_comparison/th_da_no_fill_g}
\input{exp_plots/tmaf_vs_da_comparison/th_tmaf}
\input{exp_plots/tmaf_vs_da_comparison/th_tmaf_no_fill}
\input{exp_plots/tmaf_vs_da_comparison/th_tmaf_50}
\input{exp_plots/tmaf_vs_da_comparison/th_tmaf_no_fill_50}
\input{exp_plots/tmaf_vs_da_comparison/th_tmaf_g}
\input{exp_plots/tmaf_vs_da_comparison/th_tmaf_no_fill_g}
\legend{DA~~,DA-$50\%$~V-Bat~~, DA-Gripper~~, \texttt{TMAF}~~, \texttt{TMAF}-$50\%$~V-Bat~~,  \texttt{TMAF}-Gripper}
\end{axis}
%
}
}};

\node (b) [xshift = \xshftb ex, yshift=\yshftb ex, scale=1.0]{
\scalebox{\scal}
{
\tikz{
\pgfplotsset{width=\pltw ex, height=\plth ex}
\begin{axis}[
   axis background style={fill=axisbgclr},
    title={},
    xlabel={Time (s)},
    ylabel={$e_{\text{Z}}$ (m)},
    xmin=0.0, xmax=0.99,
    ymin=-1.0, ymax=-0.25,
    xtick={0,0.99},
    xticklabels={0,20},
    extra x ticks={0, 0.20, 0.40, 0.60, 0.80, 1.00},
    extra x tick labels={},
    ytick={-1.0, -0.75, -0.50, -0.25},
    yticklabels={-$0.50$, -$0.25$, $0.0$, $0.25$},
     axis line style={axisclr},
    legend image post style={scale =\limscale},
    legend style={at={(\lxshftb ex,\lyshftb ex)},anchor=south, legend columns = 3, draw = {dlegendclr}, fill={legendclr}, nodes={scale=\lscale}},
    legend cell align={left},
    ymajorgrids=true, 
    xmajorgrids=true,
    grid style={dashed, gridclr},
    major tick length=1ex,
    x label style={at={(\xlabelxshiftb ex, \xlabelyshiftb ex)},scale=\labelscale},
    y label style={at={(\ylabelxshiftb ex, \ylabelyshiftb ex)},scale=\labelscale},
    xticklabel style={scale=\ticklabelscale},
    yticklabel style={scale=\ticklabelscale},
    legend cell align={left},
]
\FPeval{\opacty}{0.2}
\input{exp_plots/tmaf_vs_da_comparison/z_da_no_fill}
\input{exp_plots/tmaf_vs_da_comparison/z_da_no_fill_50}
\input{exp_plots/tmaf_vs_da_comparison/z_da_no_fill_g}
\input{exp_plots/tmaf_vs_da_comparison/z_tmaf_no_fill}
\input{exp_plots/tmaf_vs_da_comparison/z_tmaf_no_fill_50}
\input{exp_plots/tmaf_vs_da_comparison/z_tmaf_no_fill_g}
\legend{DA~~,DA-$50\%$~V-Bat~~, DA-Gripper~~, \texttt{TMAF}~~, \texttt{TMAF}-$50\%$~V-Bat~~,  \texttt{TMAF}-Gripper}
%
\end{axis}
%
}
}};
%
%
%
 
\end{tikzpicture}
\subfloat{\label{fig:tmafvsdath}}
\subfloat{\label{fig:tmafvsdaz}}
\vspace{-3.7ex}
\caption{Hovering performance \OURS~\textit{vs} DA \cite{direct} in different settings. Without gripper, DA achieves an RMSE of $0.16$m but only $0.03$m with TMDC. While for $50\%$ battery discharge, we observe RMSE of $0.15$m with DA and $0.02$m with \OURS. Finally, with gripper attached, DA shows an RMSE of $0.15$m and TMDC of $0.02$m.}
\label{fig:tmafvsda_hover}
\vspace{-2ex}
\end{figure}

Fig.~\ref{fig:tmafvsda_hover} shows the analysis in three different cases: (\textit{i}) flying with $100\%$ battery, (\textit{ii}) flying with $50\%$ battery, and (\textit{i}) flying with $100\%$ battery with off-center gripper weighing $900$ gms. It can be readily noticed that DA fails to maintain height when the gripper is attached to the platform whose mass is unknown to DA, and also, as battery discharges, DA does not output the required thrust, needing to adjust its $\bm \mu$ (Eq.~\ref{eq:forceda}). While our \texttt{TMAF} is free from such issues. 
\subsubsection{Disturbance Rejection }
\label{sec:distrub_tmaf_da}
\begin{figure}[t]
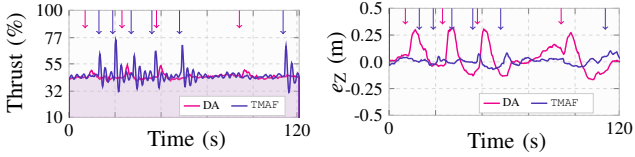

\centering

\colorlet{clr1}{white!0!magenta}
\colorlet{clr2}{blue!60!green}
\colorlet{clr3}{black!40!yellow}
\colorlet{clr4}{blue!60!brown}

\begin{tikzpicture}

\FPeval{\xshfta}{0}
\FPeval{\xshftb}{27}
\FPeval{\xshftc}{0}
\FPeval{\xshftd}{0}

\FPeval{\yshfta}{0-0}
\FPeval{\yshftb}{0-0}
\FPeval{\yshftc}{0-18}
\FPeval{\yshftd}{0-27.9}

\FPeval{\pltw}{53}
\FPeval{\plth}{30}

\FPeval{\scal}{0.45}

\FPeval{\mrksize}{0.3}

\colorlet{gridclr}{white!85!black}
\colorlet{dlegendclr}{white!80!black}
\colorlet{axisclr}{white!75!black}
\colorlet{axisbgclr}{white!99!black}

\colorlet{dlegendclr}{white!80!black}
\colorlet{legendclr}{white!100!black}
\colorlet{desclr}{white!0!brown}

\FPeval{\lscale}{1.0}
\FPeval{\limscale}{0.8}

\FPeval{\lxshfta}{30.5}
\FPeval{\lxshftb}{27.5}

\FPeval{\lyshfta}{2.5}
\FPeval{\lyshftb}{0.4}

\FPeval{\titlescale}{0.7}
\FPeval{\titlexshift}{21.0}
\FPeval{\titleyshift}{13.7}

\colorlet{gpudclr}{white!80!black}
\colorlet{gpuclr}{white!90!black}
\colorlet{gputxtclr}{white!0!black}

\FPeval{\labelscale}{1.9}
\FPeval{\ticklabelscale}{1.5}

\FPeval{\xlabelxshift}{0+22}
\FPeval{\xlabelyshift}{0+4.5}
\FPeval{\ylabelxshift}{0-2.0}
\FPeval{\ylabelyshift}{0+12.5}

\FPeval{\xlabelxshiftb}{0+22}
\FPeval{\xlabelyshiftb}{0+1.5}
\FPeval{\ylabelxshiftb}{0-2.0}
\FPeval{\ylabelyshiftb}{0+9.9}

\FPeval{\linew}{1.2}
\FPeval{\dashon}{5.0}
\FPeval{\dashoff}{3.0}

\FPeval{\xshfttakeoff}{0-16}
\FPeval{\xshfthover}{0}
\FPeval{\xshftland}{16}

\node (a) [xshift = \xshfta ex, yshift=\yshfta ex, scale=1.0]{
\scalebox{\scal}
{
\tikz{
\pgfplotsset{width=\pltw ex, height=\plth ex}
\begin{axis}[
   axis background style={fill=axisbgclr},
    title={},
    xlabel={Time (s)},
    ylabel={ Thrust ($\%$)},
    xmin=0.0, xmax=1.0,
    ymin=10, ymax=100,
    xtick={0,0.99},
    xticklabels={0,120},
    extra x ticks={0, 0.20, 0.40, 0.60, 0.80, 1.00},
    extra x tick labels={},
    ytick={10, 32, 55, 77, 100},
    yticklabels={$10$, $32$, $55$, $77$, $100$},
     axis line style={axisclr},
    legend image post style={scale =\limscale},
    legend style={at={(\lxshfta ex,\lyshfta ex)},anchor=south, legend columns = 4, draw = {dlegendclr}, fill={legendclr}, nodes={scale=\lscale}},
    legend cell align={left},
    ymajorgrids=true, 
    xmajorgrids=true,
    grid style={dashed, gridclr},
    major tick length=1ex,
    x label style={at={(\xlabelxshift ex, \xlabelyshift ex)},scale=\labelscale},
    y label style={at={(\ylabelxshift ex, \ylabelyshift ex)},scale=\labelscale},
    xticklabel style={scale=\ticklabelscale},
    yticklabel style={scale=\ticklabelscale},
]
\FPeval{\opacty}{0.2}
\input{exp_plots/disturb_rej__tmaf_vs_da/th_da}
\input{exp_plots/disturb_rej__tmaf_vs_da/th_da_no_fill}
\input{exp_plots/disturb_rej__tmaf_vs_da/th_tmaf}
\input{exp_plots/disturb_rej__tmaf_vs_da/th_tmaf_no_fill}
\legend{DA~~~, \texttt{TMAF}}
\draw [->, clr1, thick] (axis cs:0.07,100) -- (axis cs:0.07,85);
\draw [->, clr1, thick] (axis cs:0.23,100) -- (axis cs:0.23,85);
\draw [->, clr1, thick] (axis cs:0.38,100) -- (axis cs:0.38,85);
\draw [->, clr1, thick] (axis cs:0.74,100) -- (axis cs:0.74,85);
\draw [->, clr4, thick] (axis cs:0.13,100) -- (axis cs:0.13,80);
\draw [->, clr4, thick] (axis cs:0.19,100) -- (axis cs:0.19,80);
\draw [->, clr4, thick] (axis cs:0.27,100) -- (axis cs:0.27,80);
\draw [->, clr4, thick] (axis cs:0.48,100) -- (axis cs:0.48,80);
\draw [->, clr4, thick] (axis cs:0.36,100) -- (axis cs:0.36,80);
\draw [->, clr4, thick] (axis cs:0.93,100) -- (axis cs:0.93,80);
\end{axis}
%
}
}};

\node (b) [xshift = \xshftb ex, yshift=\yshftb ex, scale=1.0]{
\scalebox{\scal}
{
\tikz{
\pgfplotsset{width=\pltw ex, height=\plth ex}
\begin{axis}[
   axis background style={fill=axisbgclr},
    title={},
    xlabel={Time (s)},
    ylabel={ $e_{\text{Z}}$ (m)},
    xmin=0.0, xmax=0.99,
    ymin=-1.0, ymax=0.0,
    xtick={0,0.99},
    xticklabels={0,120},
    extra x ticks={0, 0.20, 0.40, 0.60, 0.80, 1.00},
    extra x tick labels={},
    ytick={-1.0, -0.75, -0.50, -0.25, 0.0},
    yticklabels={-$0.5$, -$0.25$, $0.0$, $0.25$, $0.5$},
     axis line style={axisclr},
    legend image post style={scale =\limscale},
    legend style={at={(\lxshftb ex,\lyshftb ex)},anchor=south, legend columns = 4, draw = {dlegendclr}, fill={legendclr}, nodes={scale=\lscale}},
    legend cell align={left},
    ymajorgrids=true, 
    xmajorgrids=true,
    grid style={dashed, gridclr},
    major tick length=1ex,
    x label style={at={(\xlabelxshiftb ex, \xlabelyshiftb ex)},scale=\labelscale},
    y label style={at={(\ylabelxshiftb ex, \ylabelyshiftb ex)},scale=\labelscale},
    xticklabel style={scale=\ticklabelscale},
    yticklabel style={scale=\ticklabelscale},
]
\FPeval{\opacty}{0.2}
\input{exp_plots/disturb_rej__tmaf_vs_da/z_da_no_fill}
\input{exp_plots/disturb_rej__tmaf_vs_da/z_tmaf_no_fill}
\legend{DA~~~, \texttt{TMAF}~~~, $z^\star$}
\draw [->, clr1, thick] (axis cs:0.07,0) -- (axis cs:0.07,-0.15);
\draw [->, clr1, thick] (axis cs:0.23,0) -- (axis cs:0.23,-0.15);
\draw [->, clr1, thick] (axis cs:0.38,0) -- (axis cs:0.38,-0.15);
\draw [->, clr1, thick] (axis cs:0.74,0) -- (axis cs:0.74,-0.15);
\draw [->, clr4, thick] (axis cs:0.13,0) -- (axis cs:0.13,-0.20);
\draw [->, clr4, thick] (axis cs:0.19,0) -- (axis cs:0.19,-0.20);
\draw [->, clr4, thick] (axis cs:0.27,0) -- (axis cs:0.27,-0.20);
\draw [->, clr4, thick] (axis cs:0.48,0) -- (axis cs:0.48,-0.20);
\draw [->, clr4, thick] (axis cs:0.36,0) -- (axis cs:0.36,-0.20);
\draw [->, clr4, thick] (axis cs:0.93,0) -- (axis cs:0.93,-0.20);
\end{axis}
%
}
}};
%
%
%
%
\end{tikzpicture}
\subfloat{\label{fig:disturb_tmafvsdath}}
\subfloat{\label{fig:disturb_tmafvsdaz}}
\vspace{-3.7ex}
\newcommand{\varrda}{\raisebox{0.1ex}{\tikz{\draw[->, clr1] (0,0) -- (0,-0.16);}}}
\newcommand{\varrtmaf}{\raisebox{0.1ex}{\tikz{\draw[->, clr2] (0,0) -- (0,-0.16);}}}
\caption{\OURS~\textit{vs} DA \cite{direct} at an external disturbance of $15$N during hovering. The arrows `\protect\varrda' and `\protect\varrtmaf' indicate disturbance introduction. In this case, DA shows a peak offset of $0.25$m whereas \OURS{} a peak offset only of $0.07$m.}
\label{fig:tmafvsda_disturb}
\vspace{-3ex}
\end{figure}

Fig.~\ref{fig:tmafvsda_disturb} shows the analysis. We introduce a sudden intensive force of $\sim15$N on the quadrotor. Noticeably, DA gets disturbed from the set point with a large variance. While \texttt{TMAF} offers stiffness against the disturbance and maintains the height by quickly adjusting the thrust. 
\begin{figure}[t]
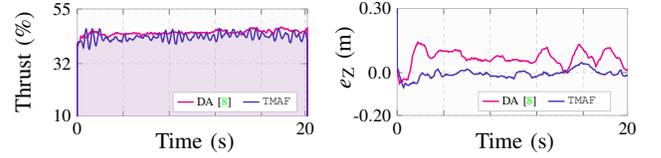

\centering

\begin{tikzpicture}

\FPeval{\xshfta}{0}
\FPeval{\xshftb}{27}
\FPeval{\xshftc}{0}
\FPeval{\xshftd}{0}

\FPeval{\yshfta}{0-0}
\FPeval{\yshftb}{0-0}
\FPeval{\yshftc}{0-18}
\FPeval{\yshftd}{0-27.9}

\FPeval{\pltw}{53}
\FPeval{\plth}{30}

\FPeval{\scal}{0.45}

\FPeval{\mrksize}{0.3}

\colorlet{clr1}{white!0!magenta}
\colorlet{clr2}{blue!60!green}
\colorlet{clr3}{black!40!yellow}
\colorlet{clr4}{blue!60!brown}

\colorlet{gridclr}{white!85!black}
\colorlet{dlegendclr}{white!80!black}
\colorlet{axisclr}{white!75!black}
\colorlet{axisbgclr}{white!99!black}

\colorlet{dlegendclr}{white!80!black}
\colorlet{legendclr}{white!100!black}
\colorlet{desclr}{white!0!brown}

\FPeval{\lscale}{1.0}
\FPeval{\limscale}{0.8}

\FPeval{\lxshfta}{29.5}
\FPeval{\lxshftb}{27.5}

\FPeval{\lyshfta}{5.2}
\FPeval{\lyshftb}{0.8}

\FPeval{\titlescale}{0.7}
\FPeval{\titlexshift}{21.0}
\FPeval{\titleyshift}{13.7}

\colorlet{gpudclr}{white!80!black}
\colorlet{gpuclr}{white!90!black}
\colorlet{gputxtclr}{white!0!black}

\FPeval{\labelscale}{1.9}
\FPeval{\ticklabelscale}{1.5}

\FPeval{\xlabelxshift}{0+22}
\FPeval{\xlabelyshift}{0+6.1}
\FPeval{\ylabelxshift}{0-2.0}
\FPeval{\ylabelyshift}{0+14.5}

\FPeval{\xlabelxshiftb}{0+22}
\FPeval{\xlabelyshiftb}{0+1.5}
\FPeval{\ylabelxshiftb}{0-2.0}
\FPeval{\ylabelyshiftb}{0+10.9}

\FPeval{\linew}{1.2}
\FPeval{\dashon}{5.0}
\FPeval{\dashoff}{3.0}

\FPeval{\xshfttakeoff}{0-16}
\FPeval{\xshfthover}{0}
\FPeval{\xshftland}{16}

\node (a) [xshift = \xshfta ex, yshift=\yshfta ex, scale=1.0]{
\scalebox{\scal}
{
\tikz{
\pgfplotsset{width=\pltw ex, height=\plth ex}
\begin{axis}[
   axis background style={fill=axisbgclr},
    title={},
    xlabel={Time (s)},
    ylabel={ Thrust ($\%$)},
    xmin=0.0, xmax=1.0,
    ymin=10, ymax=55,
    xtick={0,0.99},
    xticklabels={0,20},
    extra x ticks={0, 0.20, 0.40, 0.60, 0.80, 1.00},
    extra x tick labels={},
    ytick={10, 32, 55, 77, 100},
    yticklabels={$10$, $32$, $55$, $77$, $100$},
     axis line style={axisclr},
    legend image post style={scale =\limscale},
    legend style={at={(\lxshfta ex,\lyshfta ex)},anchor=south, legend columns = 4, draw = {dlegendclr}, fill={legendclr}, nodes={scale=\lscale}},
    legend cell align={left},
    ymajorgrids=true, 
    xmajorgrids=true,
    grid style={dashed, gridclr},
    major tick length=1ex,
    x label style={at={(\xlabelxshift ex, \xlabelyshift ex)},scale=\labelscale},
    y label style={at={(\ylabelxshift ex, \ylabelyshift ex)},scale=\labelscale},
    xticklabel style={scale=\ticklabelscale},
    yticklabel style={scale=\ticklabelscale},
]
\FPeval{\opacty}{0.2}
\input{exp_plots/ground_eff__tmaf_vs_da/th_da}
\input{exp_plots/ground_eff__tmaf_vs_da/th_da_no_fill}
\input{exp_plots/ground_eff__tmaf_vs_da/th_tmaf}
\input{exp_plots/ground_eff__tmaf_vs_da/th_tmaf_no_fill}
\legend{DA \cite{direct}~~~, \texttt{TMAF}}
\end{axis}
%
}
}};

\node (b) [xshift = \xshftb ex, yshift=\yshftb ex, scale=1.0]{
\scalebox{\scal}
{
\tikz{
\pgfplotsset{width=\pltw ex, height=\plth ex}
\begin{axis}[
   axis background style={fill=axisbgclr},
    title={},
    xlabel={Time (s)},
    ylabel={ $e_{\text{Z}}$ (m)},
    xmin=0.0, xmax=0.99,
    ymin=-0.5, ymax=0.0,
    xtick={0,0.99},
    xticklabels={0,20},
    extra x ticks={0, 0.20, 0.40, 0.60, 0.80, 1.00},
    extra x tick labels={},
    ytick={-0.50, -0.30, 0.0},
    yticklabels={-$0.20$, $0.0$, $0.30$},
     axis line style={axisclr},
    legend image post style={scale =\limscale},
    legend style={at={(\lxshftb ex,\lyshftb ex)},anchor=south, legend columns = 4, draw = {dlegendclr}, fill={legendclr}, nodes={scale=\lscale}},
    legend cell align={left},
    ymajorgrids=true, 
    xmajorgrids=true,
    grid style={dashed, gridclr},
    major tick length=1ex,
    x label style={at={(\xlabelxshiftb ex, \xlabelyshiftb ex)},scale=\labelscale},
    y label style={at={(\ylabelxshiftb ex, \ylabelyshiftb ex)},scale=\labelscale},
    xticklabel style={scale=\ticklabelscale},
    yticklabel style={scale=\ticklabelscale},
]
\FPeval{\opacty}{0.2}
\input{exp_plots/ground_eff__tmaf_vs_da/z_da_no_fill}
\input{exp_plots/ground_eff__tmaf_vs_da/z_tmaf_no_fill}
\legend{DA \cite{direct}~~~, \texttt{TMAF}~~~, $z^\star$ }
  %
\end{axis}
%
}
}};
%
%
%
 
\end{tikzpicture}
\subfloat{\label{fig:ground_tmafvsdath}}
\subfloat{\label{fig:ground_tmafvsdaz}}
\vspace{-3.7ex}
\caption{Hovering performance of \OURS~\textit{vs} DA \cite{direct} at ground effect. \OURS{} shows an RMSE offset of $0.02$m and peak offset of $0.11$m only in comparison to the significant offset of $0.12$m with DA and peak offset of $0.30$m.
}
\label{fig:tmafvsda_ground}
\vspace{-3ex}
\end{figure}
\subsubsection{Ground Effect}
\label{sec:ground_effect}
Fig.~\ref{fig:tmafvsda_ground} shows the analysis. We keep the hovering height to $0.30$m, less than the rotor diameter ($0.46$m). It can be seen that \texttt{TMAF} exhibits accurate hovering performance with  near the ground by micro-adjusting the thrust, while DA does not maintain the desired height.
%

 
%
\section{Conclusion}
   \label{sec:conc}
  In this paper, we present Thrust Microstepping and Decoupled Control (\OURS) for quadrotor targeting aerial grasping challenges, such as center and off-center dynamic payloads, battery discharge and wind drafts. \OURS~has two novel components. First is Thrust Microstepping via Acceleration Feedback (\texttt{TMAF}) thrust controller which handles unmodeled disturbances and large variances in system mass, payloads, even at low-frequency loop rates. Second is Decoupled Motion Control (\texttt{DMC}), which decouples the motion control in the horizontal and vertical directions to precisely counteract the disturbances induced during an off-center payload attachment by directly estimating desired attitude instead of using forces. \texttt{TMAF} outperforms recent direct acceleration feedback thrust controller (DA), and \texttt{DMC} outperforms geometric tracking control for attitude estimation. Overall, \OURS{} exhibits stable flights in many real-world adverse situations such as constrained flying workspace, solely onboard position feedback, limited computing power, non-uniform loop rates, etc., verified via rigorous experimental analysis.


%
\appendices
\section{}
\label{app:proof}
We prove the asymptotic stability of our thrust controller. Thus, rewriting the Eq.~\ref{eq:classicalthrust} and taking its time derivative:
%
\begin{equation}
\scriptsize
a_B = \frac{f_B}{m} + \frac{f_e}{m} - g,~~~~~~~~\dot{a}_B = \frac{\dot{f}_B}{m} + d
\end{equation}
where, $d= -\frac{\dot{m}f_B}{m^2} + \frac{\dot{f}_e}{m} - \frac{\dot{m}f_e}{m^2}$, $\dot{f}_B=\nicefrac{\Delta f_B}{T}$ using Euler's backward difference,  $T$ is the sampling time. Hence, $\dot{a}_B$ becomes $(\nicefrac{1}{mT} ) \Delta f_B + d$. Using $\dot{a}_B$ and $\dot{f}_B$ into $e_a=a_B^*-a_B$ leads to:
\begin{equation}
\scriptsize
\begin{aligned}
\dot{e}_a &=   \dot{a}^\star_B - \frac{\alpha e_a + \beta \dot{e}_a}{mT} - d =  - e_a \frac{\alpha }{\beta +mT} + \dot{a}^\star_B \frac{mT}{\beta+mT} - d \frac{mT}{\beta+mT} \\
&=-k(e_a-\bar{e}_a) \label{eq:ea}
\end{aligned}
\end{equation}
where $k=\frac{\alpha }{\beta +mT}$ and $\bar{e}_a=\dot{a}^\star_B \frac{mT}{\alpha} - d \frac{mT}{\alpha}$. It can be inferred from Eq.~\ref{eq:ea} that the acceleration error $e_a$ converges to $\bar{e}_a$. Generally, in aerial grasping, mass, external disturbance, or $\dot{a}^\star_B$ does not vary continuously and quickly after the payload attach and the detach events. Hence in the expression of $\bar{e}_a$, it can be assumed that $\dot{a}^\star_B = \dot{m} = \dot{f}_e \approx 0 $. This implies $\bar{e}_a \rightarrow 0 \implies e_a \rightarrow 0$, indicating the asymptotic stability of the thrust controller. This, in turn, results in the convergence of the position feedback PID loops (Sec.~\ref{sec:positionloop}) to the set point.

%
\bibliographystyle{ieeetr}
\bibliography{bibfile}

\end{document}